\documentclass[times,twocolumn,final,authoryear]{elsarticle}

\usepackage{template}
\usepackage{framed,multirow}

\usepackage{amssymb}
\usepackage{latexsym}

\usepackage{url}
\usepackage{xcolor}
\definecolor{newcolor}{rgb}{.8,.349,.1}

\usepackage{booktabs}

\journal{Computer Vision and Image Understanding}

\begin{document}

\begin{frontmatter}

\title{Survey and Systematization of 3D Object Detection Models and Methods}

\author[1]{Moritz Drobnitzky} 
\author[2]{Jonas Friederich}
\author[3]{Bernhard Egger}
\author[3,1]{Patrick Zschech}

\address[1]{Technische Universität Dresden, Münchner Platz 3, 01187 Dresden, Germany}
\address[2]{Mærsk Mc-Kinney Møller Institute, University of Southern Denmark, Campusvej 55, 5230 Odense, Denmark}
\address[3]{Friedrich-Alexander-Universität Erlangen-Nürnberg, Schloßplatz 4, 91054 Erlangen, Germany\\

Corresponding author: patrick.zschech@fau.de}

\received{1 May 2013}
\finalform{10 May 2013}
\accepted{13 May 2013}
\availableonline{15 May 2013}
\communicated{S. Sarkar}

\begin{abstract}

Strong demand for autonomous vehicles and the wide availability of 3D sensors are continuously fueling the proposal of novel methods for 3D object detection. In this paper, we provide a comprehensive survey of recent developments from 2012-2021 in 3D object detection covering the full pipeline from input data, over data representation and feature extraction to the actual detection modules. We introduce fundamental concepts, focus on a broad range of different approaches that have emerged over the past decade, and propose a systematization that provides a practical framework for comparing these approaches with the goal of guiding future development, evaluation and application activities. Specifically, our survey and systematization of 3D object detection models and methods can help researchers and practitioners to get a quick overview of the field by decomposing 3DOD solutions into more manageable pieces.

%} %I guess i introduced a stupid mistake in the style files
\end{abstract}


\end{frontmatter}

%\linenumbers


\section{Introduction}
\label{section:Introduction}

Gaining a high-level and three-dimensional understanding of digital pictures is one of the major challenges in the field of artificial intelligence. Applications like augmented reality, autonomous driving and other robotic navigation systems are pushing research in this field faster than ever \citep{aprile_dynamically_2008, chen_class-discriminative_2022, amirkhani_survey_2022}. Participating in real-life road traffic, self-driving vehicles need to gain an absolute understanding of their surroundings. Hence, a vehicle not only needs to recognize other road users and other objects, but also comprehend their pose and location to avoid collisions. This objective is well-known as 3D object detection (3DOD) \citep{arnold_survey_2019}.

Meanwhile, 2D object detection (2DOD) has obtained impressive results in terms of precision and inference time, and is able to compete with or even surpass human vision \citep{zhao_object_2019}. However, to fully grasp the scene in a real 3D world, 2D recognition and detection results alone are no longer sufficient. 3DOD now extends this approach into the three-dimensional space by adding the desired parameters of dimension and orientation of the object to the established location and classification results.

The literature volume for 3DOD has increased significantly over the past years \citep{fernandes_point-cloud_2021, friederich_3DOD}. Against the backdrop of highly sophisticated 2DOD models, it is apparent that the focus of research is shifting to 3DOD as the necessary hardware in terms of sensors and computing units becomes increasingly available.

Since 3DOD is a steadily growing field of investigation, there are several promising approaches and trends, including a large pool of various design options for the object detection pipeline. Providing an overview about relevant approaches and seminal achievements may offer orientation and can help to initiate further development in the research community. For this reason, we present a comprehensive review of 3DOD models and methods with exemplary applications and aim to conceptualize the full range of 3DOD approaches along a multi-stage pipeline.

With our work, we complement related surveys in the field \cite[e.g.,][]{arnold_survey_2019, guo_deep_review_2020, fernandes_point-cloud_2021}, which often focus on a particular domain (e.g., autonomous driving), specific data input (e.g., point cloud data), or a certain set of methods (e.g., deep learning techniques). 

To carry out our review, we investigated papers that were published in a period from 2012 to 2021. In total, our literature corpus comprises more than hundred papers which we examined in detail to provide a classification of all approaches. Throughout our review, we describe representative examples along the 3DOD pipeline, while highlighting seminal achievements.

This survey is structured as follows. In Section~\ref{section:foundations}, we provide all relevant foundations and subsequently refer to related work in Section~\ref{section:related_work}. Thereafter, the identified literature is discussed and analyzed in detail. This constitutes the main part of our survey, for which we propose a structured framework along the 3DOD pipeline in Section~\ref{section:3DOD-Pipeline}. The framework consists of several stages with corresponding design options, which are examined in Section~\ref{section:Input Data}-\ref{section:Detection Module} due to their thematic depth. Afterwards, we leverage our framework and classify the examined literature corpus in Section~\ref{section:Classification of literature}. Finally, we draw a conclusion of this work and give an outlook for future work in Section~\ref{section:Conclusion}.
For better readability, we include a list of all abbreviations at the end of the manuscript.

\section{Foundations}
\label{section:foundations}

To give orientation in the following sections, we introduce major concepts of computer vision and in particular of 3DOD that are regarded as background and foundational knowledge.

\subsection{Object Detection}
\label{section:Object Detection}
A core task in the field of computer vision is to recognize and classify objects in images. This general task can further be subdivided into several sub-tasks as summarized in Table {\ref{tab:Computer Vision}}.

\begin{table}[ht]
    \centering
    \begin{tabular}{l| p{45mm}}
Object recognition  & Recognize/Classify a specific \newline object type in an image \\
\hline
Object localization & Localize an object in an image                         \\
\hline
Object detection    & Localize and recognize \newline objects in an image             \\
\hline
Object segmentation & Determine which pixels belong \newline to an object in an image \\
\hline
Pose estimation     & Determine the pose of an \newline object in an image    \\ 
    \end{tabular}
    \caption[Terms of Computer Vision]{Different tasks in computer vision that focus on object instances}
    \label{tab:Computer Vision}
\end{table}

Following this distinction, object detection is the fusion of object recognition and localization. In detail, the approach tries to simultaneously classify and localize specific object instances in an image \citep{zhao_object_2019}. Detected objects are classified and usually marked with bounding boxes. These are imaginary boxes that describe the objects of interest. They are defined as the coordinates of the rectangular border that fully encloses the target object.

Object detection can be considered as a supervised learning problem which defines the process of learning a function that can map input data to known targets based on a given set of training data \citep{bishop_pattern_2006}. For this task, different kinds of machine learning algorithms can be applied.

Conventional machine learning approaches require to first extract representative image features based on feature descriptors, such as Viola-Jones method \citep{viola_robust_2004}, scale-invariant feature transform (SIFT) \citep{lowe_distinctive_2004} or histogram of oriented gradients (HOG) \citep{dalal_histograms_2005}. Those are low-level features which are manually designed for the specific use case. On their basis, prediction models such as support vector machines (SVMs) can be trained to perform the object recognition task \citep{sager_survey_2021}. However, the diversity of image objects and use cases in form of pose, illumination and background makes it difficult to manually create a robust feature descriptor that can describe all kinds of objects \citep{janiesch_2021}.

For this reason, recent efforts are increasingly directed towards the application of artificial neural networks with deep network architectures, broadly summarized under the term \textit{deep learning} \citep{lecun_deep_2015}. Deep neural networks are able to perform object recognition without having to manually define specific features in advance. Their multi-layered architecture allows them to be fed with high-dimensional raw input data and then automatically discover internal representations at different levels of abstraction that are needed for recognition and detection tasks \citep{lecun_deep_2015, bayoudh_survey_2022}.

A common type of deep neural network architecture, which is widely adopted by the computer vision community, is that of \textit{convolutional neural networks} (CNNs). Due to their nested design, they are able to process high-dimensional data that come in the form of multiple arrays, such as given by color images that are composed of arrays containing pixel intensities in different color channels \citep{liu_deep_2020}. These techniques proofed to be superior in 2DOD offering more complex and robust features, while being applicable to any use case \citep{liu_deep_2020, lecun_deep_2015}. 

\subsection{3D Vision and 3D Object Detection}
\label{section:3D Vision and 3D Object Detection}

3D vision aims to extend the previously discussed concepts of object detection by adding data of the third dimension. This leads on the one hand to six possible degrees of freedom (6DoF)\footnote{surge, heave, sway, yaw, pitch and roll} instead of three, and on the other hand, to an accompanying increase in the number of scenery configurations. While methods in 2D space are good for simple visual tasks, more sophisticated approaches are required to improve, for instance, autonomous driving or robotics applications  \citep{liang_triangulation-based_2022, liu_densernet_2021}. The full understanding of the environment composed of real 3D objects implies the interpretation of scenes in which items may show up in absolutely discretionary positions and directions related to 6DoF. This requires a substantial amount of computing power and increases the complexity of the performed operations \citep{davies_computer_2012}.

3DOD transfers the task of object detection into the three-dimensional space. The idea of 3DOD is to output dimension, location and rotation of 3D bounding boxes and the corresponding class labels for all relevant objects within the sensors field of view \citep{sager_labelcloud_2021}. 3D bounding boxes are rectangular cuboids in the three-dimensional space. To ensure relevancy, their size should be minimal, while still containing all relevant parts of an object. One common way to parameterize a 3D bounding box is $(x, y, z, h, w, l, c)$, where $(x, y, z)$ represent the 3D coordinates of the bounding box center, $(h, w, l)$ refer to the height, width and length of the box and $c$ stands for the class of the box \citep{chen_multi-view_2017}. Further, most approaches add an orientation parameter to the 3D bounding box defining the rotation of each box \cite[e.g.,][]{shi_points_2020}. 

\subsection{Sensing Technologies}
\label{section:Sensors}
To capture 3D scenes, commonly used monocular cameras are no longer sufficient. Therefore, special sensors have been developed to capture depth information. RGB-Depth (RGB-D) cameras like Intel’s RealSense use stereo vision, while Light Detection and Ranging (LiDAR) sensors such as Velodyne’s HDL-32E use laser beams to infer depth information. The data acquired by these 3D sensors can be converted to a more generic structure, the point cloud, which can be understood as a set of points in vector space. Further details on the different data inputs are provided in Section \ref{section:Input Data}.
Typically, 3DOD models rely on data captured by various active and passive optical sensor modalities, with cameras and LiDAR-sensors being the most popular representatives.

\subsubsection{Cameras}
\label{subsection:Camera}

\paragraph{Stereo cameras}

Stereo cameras are inspired by human ability to estimate depth by capturing images with two eyes. Depth gets reconstructed by exploiting the disparity between two or more camera images that record the same scene from different points of view. To do so, stereoscopy leverages triangulation and epipolar geometry theory to create range information \citep{giancola_survey_2018}. The acquired depth map normally gets appended to an RGB image as the fourth channel, together called RGB-D image. This sensor variant exhibits a dense depth map, though its quality is heavily dependent on depth estimation, which is also computationally expensive \citep{du_general_2018}.

\paragraph{Time-of-Flight cameras}

Instead of deriving depth information from different perspectives, the Time-of-Flight (TOF) principle can directly estimate the device-to-target distance. TOF systems are based on the LiDAR principle of sending light signals to the scene and measuring the time until they return. The difference between LiDAR and camera-based TOF is that LiDAR creates point clouds with a pulsed laser, whereas TOF-cameras capture depth maps with an RGB-like camera.
The data captured by stereo and TOF cameras can either be transformed into 2.5D representation like RGB-D or into 3D representation by generating a point cloud \cite[e.g.,][]{song_sliding_2014, qi_deep_2019, sun_3d_2018, ren_clouds_2020}.\\

In general, camera sensors such as stereo and TOF have the advantage of low integration costs and relatively low arithmetic complexity. However, all these methods experience considerable quality-volatility through environmental conditions like light and weather.

\subsubsection{LiDAR Sensors}
\label{subsection:LiDAR}
LiDAR sensors emit laser pulses onto the scene and measure the time from emitting the beam to receiving the reflection. In combination with the constant of the speed of light, the measured time reveals the distance to the target. By assembling the 3D spatial information from the reflected laser at a 360° angle, the sensor constructs a full three-dimensional map of the environment. This map is a set of
3D points, also called \textit{point cloud}.

The respective reflection values represent the strength of the received pulses. Thereby, LiDAR does not consider RGB \citep{lefsky_lidar_2002}. The HDL-64E3, as a common LiDAR system, outputs 120,000 points per frame, which adds up to a huge amount of data, namely 1,200,000 points per second on a 10 Hz frame rate \citep{arnold_survey_2019}.

The advantages of LiDAR sensors are long-range detection abilities, high resolution compared to other 3D sensors and independence of lighting conditions, which are counterbalanced by its high costs and bulky devices \citep{arnold_survey_2019, fernandes_point-cloud_2021}.

\subsection{Domains}
\label{section:Domain}

Looking at the extensive research on 3DOD in the past ten years, the literature can be roughly summarized into two main areas: \textit{indoor applications} and \textit{autonomous vehicle applications}. These two domains are the main drivers for the field, even though 3DOD is not strictly limited to these two specific areas, as there are also other applications conceivable and already in use, such as in retail, agriculture and fitness.

Both domains face individual challenges and opportunities which led to this differentiation.
However, it should also be noted that research in both areas is not mutually exclusive, as some 3DOD models offer solutions that are sufficiently generic and therefore do not focus on a particular domain \cite[e.g.,][]{qi_frustum_2018, xu_pointfusion:_2018, tang_transferable_2019, wang_frustum_2019-1}. 

A fundamental difference between indoor and autonomous vehicle applications is that objects in indoor environments are often stacked on top of each other. This provides opportunities for learning inter-object relationships between the target and base/carrier objects. In research, this is referred to as a holistic understanding of the inner scene, which enables a better communication between service robots and people \citep{ren_clouds_2020, huang_cooperative_2018}. The challenges with indoor applications are that scenes are often cluttered and many objects occlude each other \citep{ren_clouds_2020}.

Autonomous vehicle applications are characterized by long distances to potential objects and difficult weather conditions such as snow, rain and fog, which make the detection process more difficult \citep{arnold_survey_2019}. Objects may also occlude each other, but since objects like cars, pedestrians and traffic lights are unlikely to be on top of each other, techniques such as bird’s-eye view projection can efficiently compensate for this disadvantage \cite[e.g.,][]{beltran_birdnet:_2018, wang_fusing_2018}.

\subsection{Datasets}
\label{section:Datasets}
As seen in 2DOD, a crucial prerequisite for continous development and fast progress of algorithms is the availability of publicly accessible datasets. Extensive data is required to train learning-based state-of-the-art models. Further, they are used to apply benchmarks to compare one's own results with those of others. For instance, the availability of the dataset \textit{ImageNet} \citep{deng_imagenet} accelerated the development of 2D image classification and 2DOD models remarkably. The same phenomenon is observable in 3DOD and other tasks based on 3D data: More available data leads to a larger coverage of possible scenarios.

Similar to the domain focus, the most commonly used datasets for 3DOD developments can be roughly divided into two groups, distinguishing between autonomous driving scenarios and indoor scenes. 
In the following, we describe some of the major datasets that are publicly available. In addition, Table \ref{tab:dataset_overview} provides an overview of the described datasets including the main reference and the environment of the recordings as well as the number of scenes, frames, 3D bounding boxes and object classes.

\begin{table*}[ht]
\centering
\begin{tabular}{@{}lllllll@{}}
\toprule
Dataset & Reference & Domain/Environment & \# of scenes & \# of frames & \# of 3D boxes & \# of classes \\ \midrule
KITTI & \citet{geiger_are_2012} & Autonomous driving & 22 & 15.000 & 80.000 & 8 \\
Waymo Open & \citet{sun_scalability_2020} & Autonomous driving & 1150 & 230.000 & 12m & 4 \\
nuScenes & \citet{caesar_nuscenes_2020} & Autonomous driving & 1000 & 40.000 & 1.4m & 23 \\
NYUv2 & \citet{Silberman:ECCV12} & Indoor & 464 & 1.449 & 35.064 & 1000+ \\
SUN RGB-D & \citet{Song_2015_CVPR} & Indoor & N.A. & 10.335 & 64.500 & 800+ \\
Objectron & \citet{ahmadyan_objectron_2021} & Indoor and outdoor & 14.819 & 4m & 17.095 & 9 \\ \bottomrule
\end{tabular}
\caption{Overview of the described datasets}
\label{tab:dataset_overview}
\end{table*}

\subsubsection{Autonomous Driving Datasets}
\label{subsection:Dataset for autonomous driving}

\paragraph{KITTI}
\label{paragraph:KITTI}
The most popular dataset for autonomous driving applications is KITTI \citep{geiger_are_2012}. It consists of stereo images, LiDAR point clouds and GPS coordinates, all synchronized in time. Recorded scenes range from highways, complex urban areas and narrow country roads. The dataset can be used for various tasks such as stereo matching, visual odometry, 3D tracking and 3D object detection. For object detection, KITTI provides 7,481 training and 7,518 test frames including sensor calibration information and annotated 3D bounding boxes around the objects of interest in 22 video scenes. The annotations are categorized in easy, moderate and hard cases, depending on the object size, occlusion and truncation levels. Drawbacks of the dataset are the limited sensor configurations and light conditions: All recordings were made during daytime and mostly under sunny conditions. Moreover, the class frequencies are quite unbalanced. 75\% belong to the class car, 15\% to the class pedestrian and 4\% to the class cyclist. In natural scenarios, the missing variety challenges the evaluation of the latest methods. 

\paragraph{Waymo Open}
The Waymo Open dataset \citep{sun_scalability_2020} focuses on providing a diverse and comprehensive dataset. It consists of 1,150 videos that are exhaustively annotated with 2D and 3D bounding boxes in images and LiDAR point clouds, respectively. The data collection was conducted by using five cameras presenting a front and side view of the recording vehicle, as well as a LiDAR sensor for 360° view. Further, the data was recorded in three different cities with various light and weather conditions, providing a diverse scenery.

\paragraph{nuScenes}
NuScenes \citep{caesar_nuscenes_2020} comprises 1,000 video scenes, 20 seconds each, in the context of autonomous driving. Each scene is represented by six different camera views, LiDAR and radar data with full 360° field of view. It is significantly larger than the pioneering KITTI dataset with more than seven times as many annotations and 100 times as many images. Further, the nuScenes dataset also provides nighttime and bad weather scenarios, which is neglected in the KITTI dataset. 
On the downside, the dataset has limited LiDAR sensor quality with 34,000 points per frame and limited geographical diversity compared to the Waymo Open dataset, which covers an effective area of only five square kilometers.

\subsubsection{Indoor Datasets}
\label{subsection:Indoor Dataset}

\paragraph{NYUv2 \& SUN RGB-D}
NYUv2 \citep{Silberman:ECCV12} and its successor SUN RGB-D \citep{Song_2015_CVPR} are datasets commonly used for indoor applications. The goal of these datasets is to encourage methods focused on total scene understanding. The datasets were recorded using four different RGB-D sensors to ensure the generalizability of applied methods for different sensors. Even though SUN RGB-D inherited the 1449 labeled RGB-D frames from the NYUv2 dataset, NYUv2 is still occasionally used by nowadays methods. SUN RGB-D consists of 10,335 RGB-D images that are labelled with about 146,000 2D polygons and around 64,500 3D bounding boxes with accurate object orientation measures. Additionally, there is a room layout and scene category provided for each image. To improve image quality, short videos of every scene have been recorded. Several frames of these videos were then used to create a refined depth map.

\paragraph{Objectron}
Recently, Google released the Objectron dataset \citep{ahmadyan_objectron_2021}, which is composed of object-centric video clips capturing nine different objects categories in indoor and outdoor scenarios. The dataset consists of 14,819 annotated video clips containing over four million annotated images. Each video is accompanied by a sparse point cloud representation.

\section{Related Reviews}
\label{section:related_work}

As of today, to the best of our knowledge, there are only a limited number of reviews that aim to organize and classify the most important methods and pipelines for 3DOD.

\citet{arnold_survey_2019} were some of the first to propose a classification for 3DOD approaches with a particular focus on autonomous driving applications. Based on the input data that is passed into the detection model, they divide the approaches into (i) \textit{monocular image-based methods}, (ii) \textit{point cloud methods} and (iii) \textit{fusion-based methods}. Furthermore, they break down the point cloud category into three subcategories of data representation: (ii-a) \textit{projection-based}, (ii-b) \textit{volumetric representations} and (ii-c) \textit{Point Nets}. The data representation states which kind of input the model consumes and which information the input contains so that the subsequent stage can process it more conveniently according to the design choice.

While regarding various applications, such as 3D object classification, semantic segmentation and 3DOD, \citet{liu_deep_2019} focus on feature extraction methods which constitutes the properties and characteristics that the model derives from the passed data. They classify deep learning models on point clouds into (i) \textit{point-based methods} and (ii) \textit{tree-based methods}. The former directly uses the raw point cloud and the latter first employs a $k$-dimensional tree to preprocess the corresponding data representation. 

\citet{griffiths_review_2019} consider object detection as a special type of classification and thus provide relevant information for 3DOD in their review on deep learning techniques for 3D sensed data classification. They differentiate the approaches on behalf of the data representation into (i) \textit{RGB-D methods\textit}, (ii) \textit{volumetric approaches}, (iii) \textit{multi view CNNs},  (iv) \textit{unordered point set processing methods} and (v) \textit{ordered point set processing techniques}.

\citet{huang_survey_2020} touch lightly upon 3DOD in their review paper about autonomous driving technologies using deep learning methods. They suggest a similar classification of methods as \citet{arnold_survey_2019} by distinguishing between (i)~\textit{camera-based methods}, (ii)~\textit{LiDAR-based methods}, (iii)~\textit{sensor-fusion methods} and additionally (iv)~\textit{radar-based methods}. 
While giving a coarse structure for 3DOD, the conference paper is waiving an explanation for their classification. 

\citet{bello_review_2020} consider the field from a broader perspective by providing a survey of deep learning methods on 3D point clouds. The authors organize and compare different methods based on a structure that is task-independent. Subsequently, they discuss the application of exemplary approaches for different 3D vision tasks, including classification, segmentation and object detection.

Addressing likewise the higher-level topic of deep learning for 3D point clouds, \citet{guo_deep_review_2020} give a more detailed look into 3DOD. They structure the approaches for handling point clouds into (i)~\textit{region proposal-based methods}, (ii) \textit{single shot methods} and (iii)~\textit{other methods} by categorizing them on account of their model design choice. Additionally, the region proposal-based methods are split along their data representation into (i-a)~\textit{multi-view},  (i-b)~\textit{segmentation},  (i-c)~\textit{frustum-based} and again  (i-d) \textit{other methods}. Likewise, the single shot category inherits the subcategories  (ii-a)~\textit{bird’s-eye view}, (ii-b)~\textit{discretization} and (ii-c)~\textit{point-based} approaches. 

Most recently, \citet{fernandes_point-cloud_2021} presented a comprehensive survey which might be the most similar to this work. They developed a detailed taxonomy for point-cloud-based 3DOD. In general, they divide the detection models along their pipeline into three stages, namely \textit{data representation}, \textit{feature extraction} and \textit{detection network modules}.

The authors note that in terms of data representation, the existing literature takes the approach of either converting the point cloud data into voxels, pillars, frustums or 2D-projections, or consuming the raw point cloud directly.
Feature extraction gets emphasized as the most crucial part of the 3DOD pipeline. Suitable features are essential for an optimal feature learning which in turn has a great impact on the appropriate object localization and classification in later steps. The authors classify the extraction methods into point-wise, segment-wise, object-wise and CNNs which are further divided into 2D CNN and 3D CNN backbones.
The detection network module consists of the multiple output task of object localization and classification, as well as the regression of 3D bounding box parameters and object orientation. Same as in 2DOD, these modules are categorized into the architectural design principles of single-stage and dual-stage detectors.

Although all preceding reviews provide some  systematization for 3DOD, they move – with exception of \citet{fernandes_point-cloud_2021} - on a high level of abstraction. They tend to lose some of the information which is crucial to fully map relevant trends in this vivid research field.

Moreover, as mentioned above, all surveys are limited to either domain-specific aspects (e.g., autonomous driving applications) or focus on a subset of methods (e.g., point cloud-based approaches). Monocular-based methods, for example, are neglected in almost all existing review papers.

\section{3D Object Detection Pipeline}
\label{section:3DOD-Pipeline}

Intending to structure the research field of 3DOD from a broad perspective, we propose a systematization that enables to classify current 3DOD approaches at an appropriate abstraction level, by neither losing relevant information caused by a high level of abstraction nor being too specific and complex by a too fine-granular perspective. Likewise, our systematization aims at being sufficiently robust to allow a classification of all existing 3DOD pipelines and methods as well as of future works without the need for major adjustments to the general framework.

\begin{figure*}[htbp]
\centering
\includegraphics[width=0.8\textwidth]{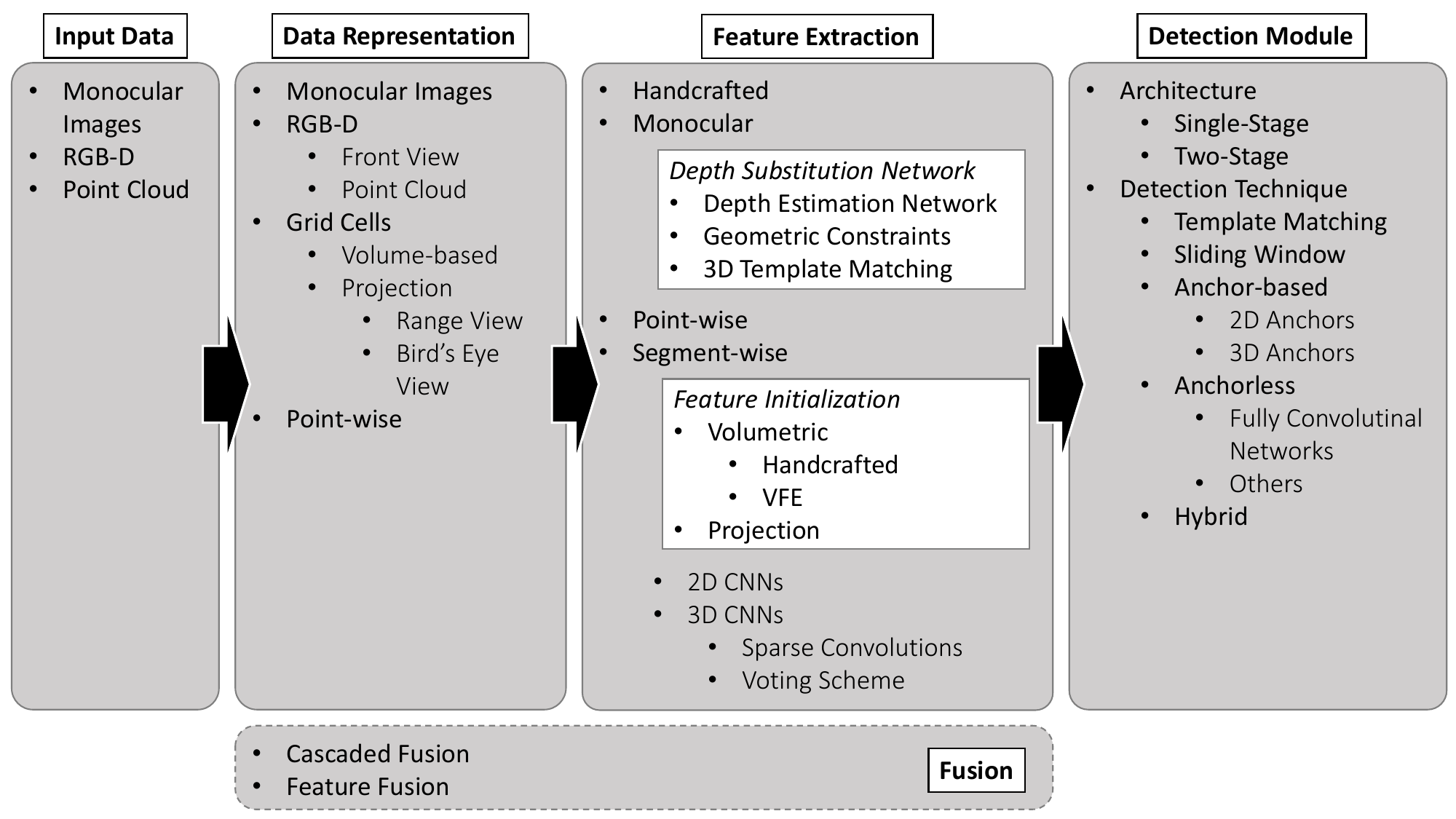}
\caption[Systematization of 3D object detection] {Systematization of a 3D object detection pipeline with its individual fine-branched design choices\label{fig:3DOD pipeline}}
\end{figure*}

Figure \ref{fig:3DOD pipeline} provides an overview of our systematization. It is structured along the general stages of an object detection pipeline, with several design choices at each stage. It starts with the choice of \textit{input data} (Section~\ref{section:Input Data}), followed by the selection of a suitable \textit{data representation} (Section~\ref{section:Data Representation}) and corresponding approaches for \textit{feature extraction} (Section~\ref{section:Feature Extraction}). For the latter steps, it is possible to apply \textit{fusion approaches} (Section~\ref{section:Fusion}) to combine different data inputs and take advantage of multiple feature representations. Finally, the \textit{object detection module} is defined (Section~\ref{section:Detection Module}).

The structuring along the pipeline enables us to order and understand the underlying principles of this field. Furthermore, we can compare the different approaches and are able to outline research trends in different stages of the pipeline. To this end, we carry out a qualitative literature analysis of proposed 3DOD approaches in the following sections along the pipeline to examine specific design options, benefits, limitations and trends within each stage.

\section{Input Data}
\label{section:Input Data}

In the first stage of the pipeline, a model consumes the input data which already restricts the further processing. Common inputs for 3DOD pipelines are (i)~\textit{RGB images} (Section~\ref{section:Input_Monocular}), (ii)~\textit{RGB-D images} (Section~\ref{section:Input_RGB-D}) and (iii)~\textit{point clouds} (Section~\ref{section:Input_Pointcloud}). 3DOD models using RGB-D images are often referred to as 2.5D approaches \cite[e.g.,][]{deng_amodal_2017, sun_3d_2018, maisano_reducing_2018}, whereas 3DOD models using point clouds are regarded as true 3D approaches.

\subsection{RGB Images}
\label{section:Input_Monocular}
Monocular or RGB images provide a dense pixel representation in the form of texture and shape information \citep{liang_deep_2018, giancola_survey_2018, arnold_survey_2019}. A 2D image can be seen as a matrix, containing the dimensions of height and width with the corresponding color values.

Especially for subtasks of 3DOD applications such as lane line detection, traffic light recognition or object classification, monocular based approaches enjoy the advantage of real time processing by 2DOD models. Probably the most severe disadvantage of monocular images is the lack of depth information. 3DOD benchmarks have shown that depth data is essential for accuracte 3D localization \citep{kitti_benchmark3DOD}. Furthermore, monocular images face the problem of object occlusion as they only capture a single view.

\subsection{RGB-D Images}
\label{section:Input_RGB-D}
RGB-D images can be created with stereo or TOF cameras that provide depth information in addition to color information (cf. Section~\ref{section:Sensors}). RGB-D images consist of an RGB image with an additional depth map \citep{du_general_2018}. The depth map is comparable to a grayscale image, except that each pixel represents the actual distance between the sensor and the surface of the scene object. An RGB image and a depth image ideally have a one-to-one correspondence between pixels \citep{wang_overview_2020}.

RGB-D images, also known as range images, are convenient to use with the majority of 2DOD methods, treating depth information similarly to the three RGB channels \citep{giancola_survey_2018}. However, as with monocular images, RGB-D faces the problem of occlusion since the scene is only presented through a single perspective. In addition, objects are presented at different scales depending on their position in space.

\subsection{Point Cloud}
\label{section:Input_Pointcloud}
The data acquired by 3D sensors can be converted to a more generic structure, the point cloud. It is a three-dimensional set of points that has an unorganized spatial structure \citep{otepka_georeferenced_2013}. The point cloud is defined by its points, which comprise the spatial coordinates of a sampled surface of an object. However, further geometric and visual attributes can be added to each point \citep{giancola_survey_2018}.

As described in Section~\ref{section:Sensors}, point clouds can be obtained from LiDAR sensors or transformed RGB-D images. Yet, point clouds obtained from RGB-D images are typically noisier and sparser compared to LiDAR-generated point clouds due to low resolution and perspective occlusion \citep{luo_3d-ssd_2020}.

The point cloud offers a fully three-dimensional reconstruction of the scene, providing rich geometric, shape and scale information. This enables the extraction of meaningful features that boost the detection performance. Nevertheless, point clouds face severe challenges which are based on their nature and processability. Common deep learning operations, which have proven to be the most effective techniques for object detection, require data to be organized in a tensor with a dense structure (e.g., images, videos) which is not fulfilled by point clouds \citep{zhou_voxelnet_2018}. In particular, point clouds exhibit \textit{irregular}, \textit{unstructured} and \textit{unordered} data characteristics \citep{bello_review_2020}.

\begin{itemize}
    \item \textit{Irregular} means that the points of a point cloud are not evenly sampled across the scene. Hence, some of the regions have a denser distribution of points than others. Especially distant objects are usually represented sparsely by very few points because of the limited range recording ability of current sensors.
    \item \textit{Unstructured} means that points are not on a regular grid. Accordingly, the distances between neighboring points can vary. In contrast, pixels in an image always have a fixed position to their neighbors, which is evenly spaced throughout the image.
    \item \textit{Unordered} means that the point cloud is just a set of points that is \textit{invariant to permutations} of its members. Particularly, the order in which the points are stored does not change the scene that it represents. In other formats, e.g. an image, data usually gets stored as a list \citep{qi_pointnet_2017, bello_review_2020}. Permutation invariance, however, means that a point cloud of $N$ points has $N!$ permutations and the subsequent data processing must be invariant to each of these different representations. 
\end{itemize}

\begin{figure*}[!htb]
\minipage{0.38\textwidth}
  \includegraphics[width=\linewidth]{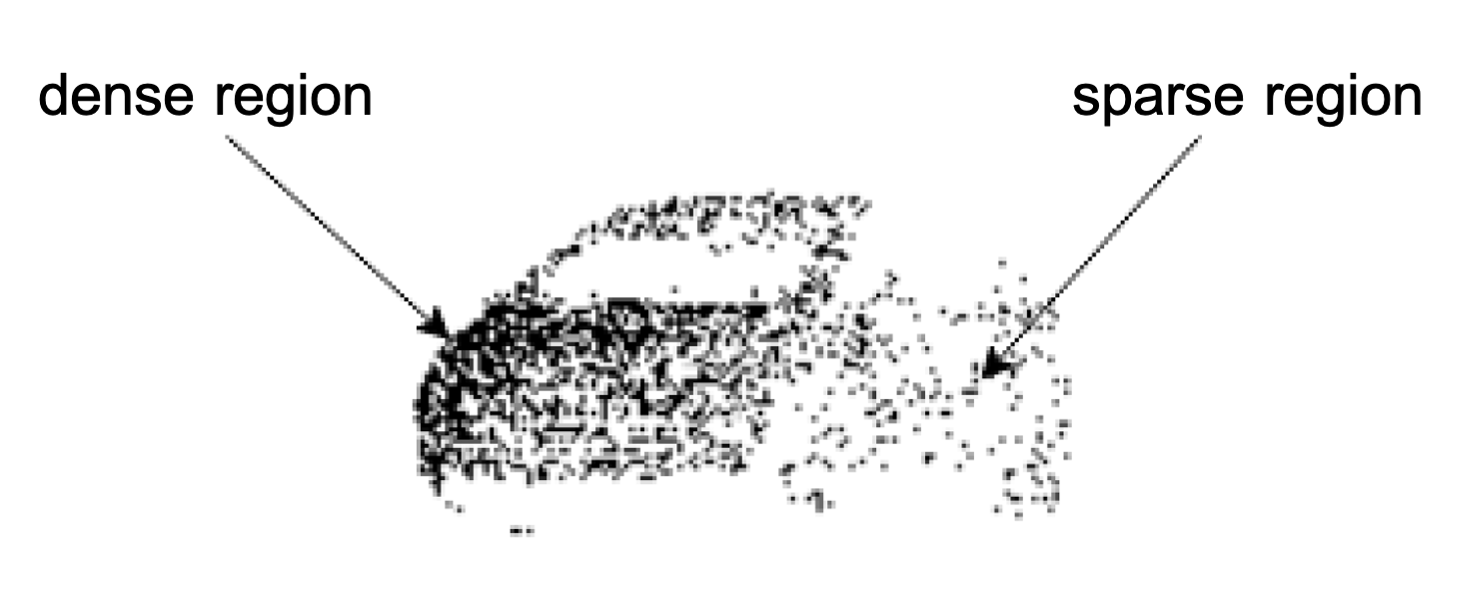}
  \centering{\textbf{(a)}}
  %\caption{Irregularity}\label{fig:awesome_image1}
\endminipage\hfill
\minipage{0.15\textwidth}
  \includegraphics[width=\linewidth]{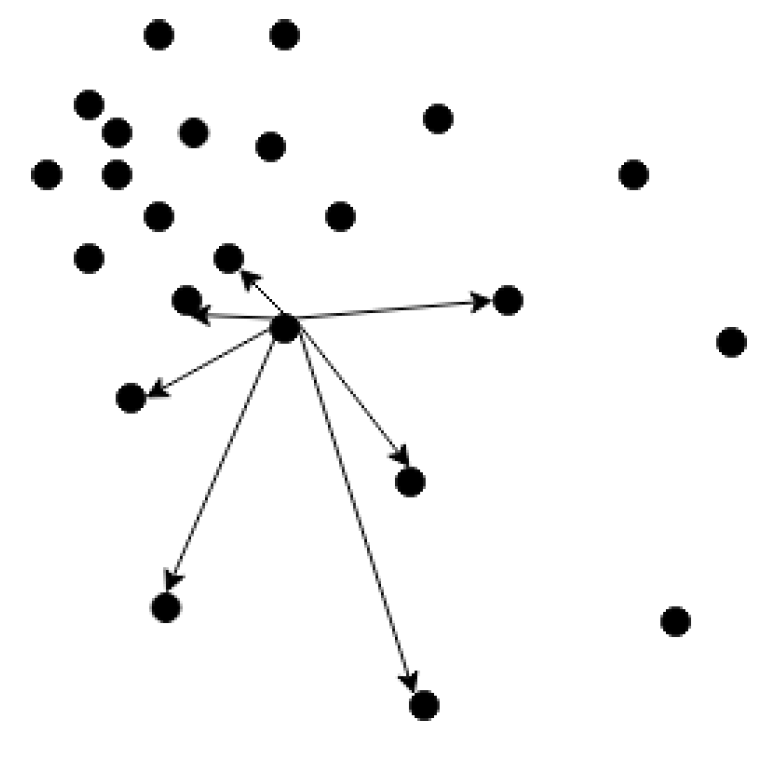}
  \centering{\textbf{(b)}}
  %\caption{Unstructured}\label{fig:awesome_image2}
\endminipage\hfill
\minipage{0.38\textwidth}%
  \includegraphics[width=\linewidth]{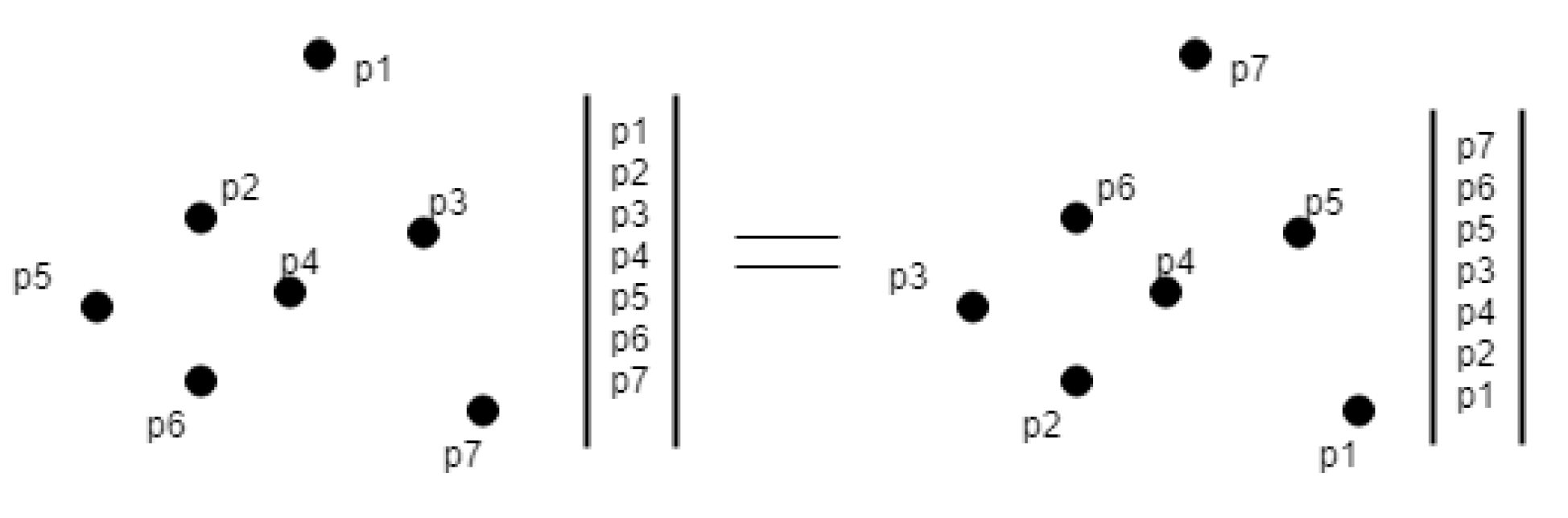}
  \centering{\textbf{(c)}}
  %\caption{Permutation invariance}\label{fig:awesome_image3}
\endminipage
\caption{Characteristics of point cloud data:  (a) Irregular collection with sparse and dense regions, (b) Unstructured cloud of independent points without a fixed grid, (c)~Unordered set of points that are invariant to permutation \citep{bello_review_2020}  }\label{fig:point_cloud_challenges}
\end{figure*}

Figure \ref{fig:point_cloud_challenges} provides an illustrative overview of three challenging characteristics of point cloud data.

\section{Data Representation}
\label{section:Data Representation}

To ensure a correct processing of the input data by 3DOD models, it must be available in a suitable representation. Due to the different data formats, 3DOD data representations can be generally classified into 2D and 3D representations. Beyond that, we assign 2.5 representations to either 2D if they come in an image format (regardless of the number of channels), or 3D if the data gets described in a spatial structure.
2D representations generally cover (i)~\textit{monocular representations} (Section~\ref{section:Data Representation_Monocular}) and (ii)~\textit{RGB-D front views} (Section~\ref{section:Data Representation_RGB-d}). 3D representations cover (iii)~\textit{grid cells} (Section~\ref{section:Data Representation_Gridcells}) and (iv)~\textit{point-wise representations} (Section~\ref{section:Data Representation_Point-Based}).

\subsection{Monocular Representation}
\label{section:Data Representation_Monocular}
Despite the lack of information about the range, monocular representation enjoys a certain popularity among 3DOD methods due to its efficient computation. Additionally, the required cameras are affordable and simple to set up. Hence, monocular representations are attractive for applications where resources are limited \citep{ku_monocular_2019, jorgensen_monocular_2019}.

The vast majority of monocular representations use the widely-known frontal view which is limited by the viewing angle of the camera. Other than that, \cite{payen_de_la_garanderie_eliminating_2018} tackle monocular 360° panoramic imagery using equirectangular projections instead of rectilinear projections of conventional camera images. To access true 360° processing, they fold the panorama imagery into a 360° ring by stitching left and right edges together with a 2D convolutional padding operation.

Only a small proportion of 3DOD monocular approaches exclusively use an image representation \cite[e.g.,][]{jorgensen_monocular_2019} for 3D spatial estimations. Most models leverage additional data and information to substitute the missing depth information (more details in section \ref{section:Feature Extraction_Deep Learning_Informed Mono}). Additionally, representation fusion techniques are quite popular to compensate disadvantages. For instance, 2D candidates get initially detected from monocular images before a 3D bounding box for the spatial object is predicted based on the initial proposals. In general, the latter step processes an extruded 3D subspace derived from the 2D bounding box. In case of representation fusion, the monocular representation is usually not used for full 3DOD but rather as a support for increasing efficiency through limiting the search space for heavy three-dimensional computations or for delivering additional features such as texture and color. These methods are described in depth in Section \ref{section:Fusion} (Fusion Approaches). 

\subsection{RGB-D Front View}
\label{section:Data Representation_RGB-d}
RGB-D data can either be transformed to a point cloud or kept in its natural form of four channels. Therefore, we can distinguish between an \textit{RGB-D (3D)} representation \cite[e.g.,][]{chen_3d_2018, tang_transferable_2019, ferguson_2d-3d_2019}, which exploits the depth information in its spatial form of a point cloud, and an \textit{RGB-D (2D)} representation \cite[e.g.,][]{chen_3d_2015, he_3d_2017, li_stereo_2019, rahman_3d_2019, luo_3d-ssd_2020}, which holds an additional 2D depth map in an image format.

Thus, as mentioned in section \ref{section:Input_RGB-D}, RGB-D (2D) images represent monocular images with an appended fourth channel of the depth map. The data is compressed along the $z$-axis generating a dense projection in the frontal view. The 2D depth image can be processed similarly to RGB channels by common 2D CNN models.

The RGB-D (2D) representation is often referred to as \textit{front view} (FV) in 3DOD research. However, front and \textit{range view} (RV) are occasionally equated with each other in current research. 
For clarification: In this work, the FV is considered as an RGB-D image generated by a TOF, stereo or similar camera, while the RV is defined as the natural frontal projection of a point cloud (see also Section \ref{section:Data Representation_projection}).

\subsection{Grid Cells}
\label{section:Data Representation_Gridcells}
Processing a point cloud (cf. Section \ref{section:Input_Pointcloud}) poses a particular challenge for CNNs because convolution operations require a structured grid, which is not present in point cloud data \citep{bello_review_2020}. Thus, to tak advantage of advanced deep learning methods and leverage highly informative point clouds, they must first be transformed into a suitable representation.

Current research presents two ways of handing point clouds. The first and more natural solution is to fit a regular grid onto the point cloud, producing a grid cell representation. Many approaches do so by either quantizing point clouds into 3D \textit{volumetric grids} (Section~\ref{section:Data Representation_volume_based}) \cite[e.g.,][]{song_sliding_2014, zhou_voxelnet_2018, shi_points_2020} or by discretizing them into \textit{(multi-view) projections} (Section~\ref{section:Data Representation_projection}) \cite[e.g.,][]{li_vehicle_2016, chen_multi-view_2017, beltran_birdnet:_2018, zheng_cia-ssd_2020}.

The second and more abstract way to solve the point cloud representation problem is to process the point cloud directly by grouping points into point sets. This approach does not require convolutions and thus allows the point cloud to be processed without transforming it into a \textit{point-wise representation} (Section~\ref{section:Data Representation_Point-Based}) \cite[e.g.,][]{qi_frustum_2018, shi_pointrcnn_2019, huang_epnet_2020}.

Along these directions, several state-of-the-art methods have been proposed for 3DOD pipelines, which we will describe exemplarily in the following.

\subsubsection{Volumetric Grids}
\label{section:Data Representation_volume_based}
\begin{figure}[htbp]
\centering
\includegraphics[width=1\columnwidth]{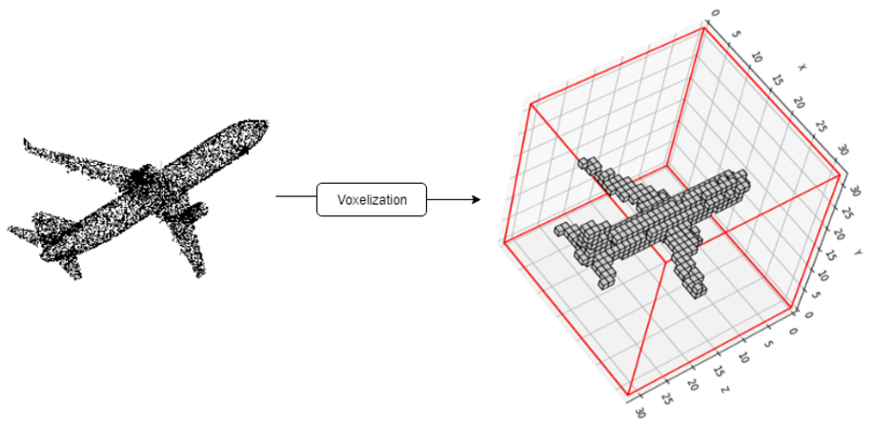}
\caption[Voxelization of Point Cloud] {Voxelization of a point cloud \citep{bello_review_2020}\label{fig:voxelization}}
\end{figure}

The main idea behind the volumetric representation is to subdivide the point cloud into equally distributed grid cells, called voxels, which allows further processing in a structured form. For this purpose, the point cloud is converted into a 3D fixed-size voxel structure of dimension $(x, y, z)$. The resulting voxels either contain raw points or already encode the occupied points into a feature representation such as point density or intensity per voxel \citep{bello_review_2020}. Figure \ref{fig:voxelization} illustrates the transformation of a point cloud into a voxel-based representation.

Usually, voxels have a cuboid shape \cite[e.g.,][]{li_3d_2017, zhou_voxelnet_2018, ren_clouds_2020}. However, there are also approaches applying other forms such as pillars \cite[e.g.,][]{lang_pointpillars_2019, lehner_patch_2019}.

Feature extraction networks utilizing voxel-based representations are computationally more efficient and reduce memory needs. Instead of extracting low- and high-dimensional features for each point individually, clusters of points (i.e., voxels) are used to extract such features \citep{fernandes_point-cloud_2021}. Despite reducing the dimensionality of a point cloud through discretization, the spatial structure is kept and allows to make use of the geometric features of the scene.

Volumetric approaches using cuboid transformation of the point cloud scene have been used, for example, by \citet{song_sliding_2014}, \citet{li_3d_2017}, \citet{engelcke_vote3deep_2017}, \citet{zhou_voxelnet_2018} and \citet{ren_clouds_2020}.

To speed up computation, \citet{lang_pointpillars_2019} propose a pillar-based voxelization of the point cloud, instead of using the conventional cubical quantization. The vertical column representation allows to skip expensive 3D convolutions in the following steps, since pillars have unlimited spatial extent in $z$-direction and can therefore be projected directly onto 2D pseudo images. As a result, all feature extraction operations are processable by efficient 2D CNNs. \citet{kuang_voxel-fpn_2020} extent this approach even further by using a learning-based feature encoding approach as opposed to relying on handcrafted feature initialization.

\subsubsection{Projection-based Representation}
\label{section:Data Representation_projection}
In addition to the need to transform the point cloud into a processable state, several approaches seek to leverage the expertise and power of 2DOD processing. Especially for reasons of better inference times for point cloud models, projection approaches became popular. They project the point cloud onto an image plane whilst preserving the depth information. Subsequently, the representation can be processed by efficient 2D extractors and detectors. Commonly used representations are the previously mentioned \textit{range view}, using an image plane projection, and the \textit{bird's eye view}, projecting the point cloud onto the ground plane. 

\paragraph{Range View}
%\label{section:Data Representation_range view}
Whereas the 2D FV corresponds to monocular and stereo cameras, the RV is a native 2D representation of the LiDAR data. The point cloud is projected onto a cylindrical 360° panoramic plane exactly as the data is captured by the LiDAR sensor. Since the LiDAR projection is still neither in a processable state nor contains any discriminative features such as RGB information, the projected RV is partitioned into a fine-grained grid and encoded in the successive feature initialization step (see Section \ref{section:Feature Extraction_Feature Initalization}).

\citet{meyer_lasernet_2019} and \citet{liang_rangercnn_2020} emphasize that the naturally compact RV results in a more efficient computation in comparison to other projections. In addition, the information loss of projection is considerably small since the RV constitutes the native representation of a rotating LiDAR sensor \citep{liang_rangercnn_2020}.
At the same time, RV suffers from distorted object size and shape due to its cylindrical image character \citep{yang_pixor_2018}. Inevitably, RV representations face the same problem as camera images, in that the size of objects is closely related to their range and occlusions may occur due to perspective \citep{zhou_fvnet_2019}. \citet[][p.2]{liang_rangercnn_2020} argue that \textit{“range image is a good choice for extracting initial features, but not a good choice for generating anchors”}. Moreover, its performance in 3DOD models does not match with state-of-the-art \textit{bird's eye view} (BEV) projections. Nevertheless, RV enables more accurate detection of small objects \citep{meyer_sensor_2019, meyer_lasernet_2019}.

Exemplary approaches using RV are proposed by \citet{li_vehicle_2016}, \citet{chen_multi-view_2017}, \citet{zhou_fvnet_2019} and \citet{liang_rangercnn_2020}.

\begin{figure}[htbp]
\centering
\includegraphics[width=1\columnwidth]{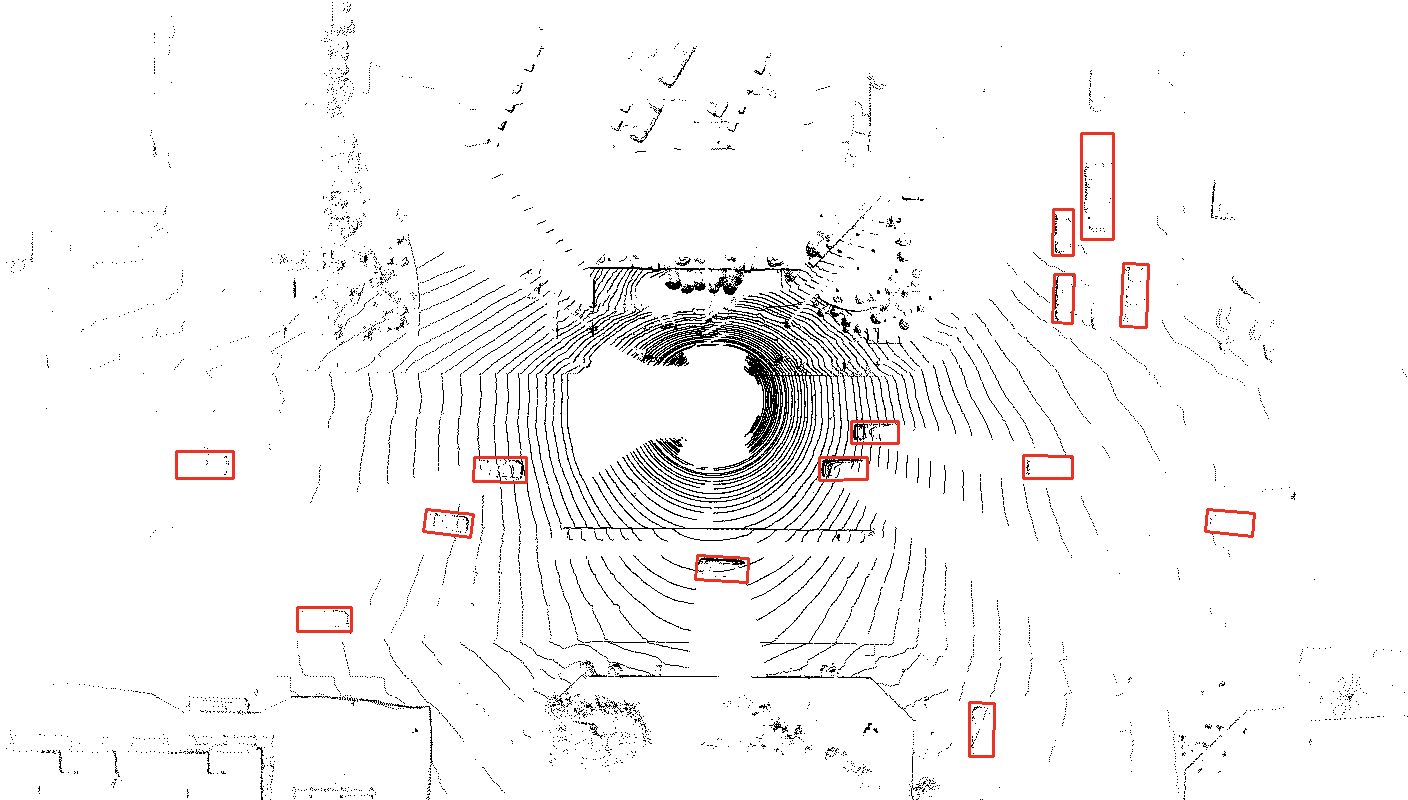}
\caption[Bird's Eye View Representation] {Visualization of bird's eye view projection of a point cloud without discretization into grid cells \citep{yang_hdnet_2018}\label{fig:BEV}}
\end{figure}

\paragraph{Bird's Eye View}
%\label{section:Data Representation_BEV}
While RV discretize the data onto a panoramic image plane, the BEV is an orthographic view of the point cloud scene projecting the points onto the ground plane. Therefore, data gets condensed along the $y$-axis. Figure \ref{fig:BEV} shows an exemplary illustration of a BEV projection.

\citet{chen_multi-view_2017} were some of the first introducing BEV to 3DOD in their seminal work MV3D. They organized the point cloud as a set of voxels and then transformed each voxel column through an overhead perspective into a 2D grid with a specific resolution and encoding for each cell. As a result, a dense pseudo-image of the ground plane is generated which can be processed by standard image detection architectures.

Unlike RV, BEV offers the advantage that the object scales are preserved regardless of the range. Further, BEV perspective eases the typical occlusion problem of object detection since objects are displayed separately from each other in a free-standing position~\citep{zhou_fvnet_2019}. These advantages let the networks exploit priors about the physical dimension of objects, especially for anchor generation \citep{yang_pixor_2018}.

On the other hand, BEV data gets sparse in at distance, which makes it unfavorable for small objects \citep{xu_pointfusion:_2018,lang_pointpillars_2019}. Furthermore, the assumption that all objects lie on one mutual ground plane often turns out to be infeasible in reality, especially in indoor scenarios \citep{zhou_fvnet_2019}. Also the often coarse voxelization of BEV may remove fine granular information leading to inferior detection at small object sizes \citep{meyer_sensor_2019}.

Exemplary models using BEV representation can be found in the work from \citet{wang_fusing_2018}, \citet{beltran_birdnet:_2018}, \citet{liang_deep_2018}, \citet{yang_pixor_2018}, \citet{simon_complex-yolo_2019},
\citet{zeng_rt3d_2018}, \citet{li_one-stage_2019_hu}, \citet{ali_yolo3d_2019}, \citet{he_structure_2020}, \citet{liang_rangercnn_2020} and \citet{wang_multi-view_2020}.

\paragraph{Multi-View Representation}
%\label{section:Data Representation_MV Representation}
Often BEV and RV are not used as single data representations but as a multi-view approach, meaning that RV and BEV but also monocular-based images are combined to represent the spatial information of a point cloud.

\citet{chen_multi-view_2017} were the first to integrate this concept into a 3DOD pipeline, followed by many other models adapting to fuse 2D representations from different perspectives \cite[e.g.,][]{ku_joint_2018, li_one-stage_2019_hu, wang_multi-channel_2019, wang_multi-view_2020, liang_rangercnn_2020}.

Although the representations of BEV and RV are compact and efficient, they are always limited by the loss of information, originating from the discretization of the point cloud into a fixed number of grid-cells and the respective feature encoding of the cells’ points. 

\subsection{Point-wise Representation}
\label{section:Data Representation_Point-Based}
Either way, discretizing the point cloud into a projection or volumetric representation inevitably leads to information loss. Against this backdrop, \citet{qi_pointnet_2017} introduced \textit{PointNet}, and thus a new way to consume the raw point cloud in its unstructured nature having access to all the recorded information.

In point-wise representations, points are isolated and sparsely distributed in a spatial structure representing the visible surface, while preserving precise localization information. PointNet handles this representation by aggregating neighboring points and extracting a compressed feature from the low-dimensional point features of each set, enabling a raw point-based representation for 3DOD models. A more detailed description of PointNet and its successor \textit{PointNet++} is given in Sections~\ref{section:Feature Extraction_PointNet} and \ref{section:Feature Extraction_PointNet++}, respectively.

However, PointNet was developed and tested on point clouds containing 1,024 points \citep{qi_pointnet_2017}, whereas realistic point clouds captured by a standard LiDAR sensor such as Velodyne's HDL-64E3 usually consist of 120,000 points per frame. Thus, applying PointNet on the whole point cloud is a time- and memory-consuming operation. De facto, point clouds are rarely consumed in total. As a consequence, further techniques are required to improve efficiency, such as cascading fusion approaches (see Section \ref{section:Fusion_Cascaded Fusion}) that crop point clouds to the region of interest and pass only subsets of the point cloud to the point-based feature extraction stage.

In general, it can be stated that point-based representations retain more information than voxel- or projection-based methods. But on the downside point-based methods are inefficient when the number of points is large. Yet, a reduction of the point clouds like in cascading fusion approaches always comes with a decrease in information. In summary, it can be said that maintaining both efficiency and performance is not achievable for any of the representations to date.

\section{Feature Extraction}
\label{section:Feature Extraction}

Feature extraction gets emphasized as the most crucial part of the 3DOD pipeline that research is focusing on \citep{fernandes_point-cloud_2021}. It follows the paradigm to reduce dimensionality of the data representation with the intention of representing the scene by a robust set of features. Features generally depict the unique characteristics of the data used to bridge the semantic gap, which denotes the difference between the human comprehension of the scene and the model’s prediction. 
Suitable features are essential for an optimal feature learning which in turn has a great impact on the detection in later steps. Hence, the goal of feature extraction is to provide a robust semantic representation of the visual scene that ultimately leads to the recognition and detection of different objects \citep{zhao_object_2019}. 

As with 2DOD, feature extraction approaches can be roughly divided into (i) \textit{handcrafted feature extraction} (Section~\ref{section:Feature Extraction_Handcrafted}) and (ii) \textit{feature learning} via deep learning methods. Regarding the latter, we can distinguish the broad body of 3DOD research depending on the respective data representation. Hence, feature learning can be performed either in a (ii-a)~\textit{monocular} (Section~\ref{section:Feature Extraction_Deep Learning_Mono}), (ii-b)~\textit{point-wise} (Section~\ref{section:Feature Extraction_Pointwise}), (ii-c)~\textit{segment-wise} (Section~\ref{section:Feature Extraction_Segmentwise}) or in a (ii-d) \textit{fusion-based} approach (Section~\ref{section:Fusion}).

\subsection{Handcrafted Feature Extraction}
\label{section:Feature Extraction_Handcrafted}
Although the vast majority of 3DOD approaches have moved towards hierarchical deep learning, which can generate more complex and robust features, there are still cases where features are created manually.

Handcrafted feature extraction differs from feature learning in that the features are selected individually and are usually used directly for the final determination of the scene. Features like edges or corners are tailored by hand and serve as the ultimate characteristics for object detection. There is no algorithm that independently learns how these features are constructed or how they can be combined, as it is the case with CNNs. The feature initialization already represents the feature extraction step. Often, these handcrafted features are then scored by SVMs or random forest classifiers, which are deployed exhaustively over the entire image or scene.

\begin{figure}[htbp]
\centering
\includegraphics[width=1\columnwidth]{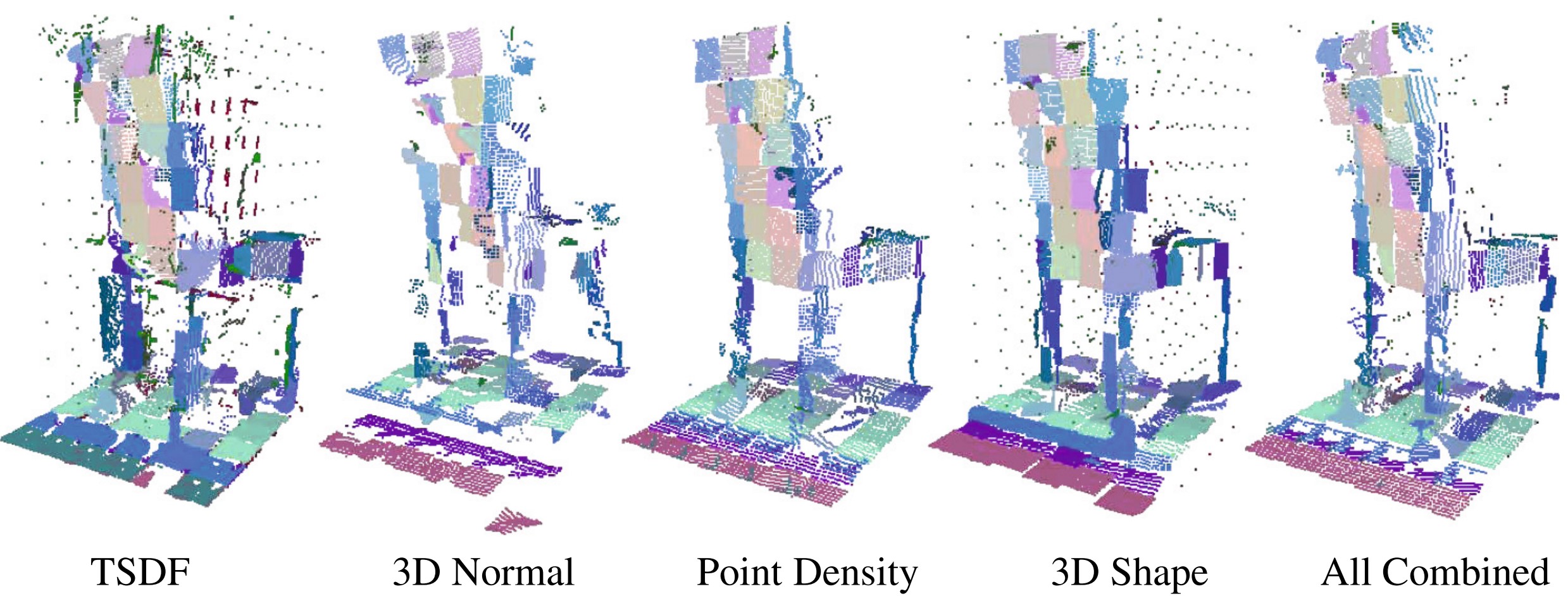}
\caption[Visualization of Handcrafted Features] {Visualization of manually crafted features \citep{song_sliding_2014}\label{fig:handcrafted features}}
\end{figure}

A few exemplary 3DOD models using handcrafted feature extraction shall be introduced in the following.
For instance, \citet{song_sliding_2014} use four types of 3D features which they exhaustively extract from each voxel cell, namely point density, 3D shape feature, 3D normal feature and truncated signed distance function feature. The features are used to handle the problem of self-occlusion (see Figure~\ref{fig:handcrafted features}).

\citet{wang_voting_2015} also use a fixed-dimensional feature vector containing the mean and variance of the reflectance values of all points that lie within a voxel and an additional binary occupancy feature. The features are not processed further and are used directly for detection purposes in a voting scheme.

\citet{ren_three-dimensional_2016} introduce their discriminative cloud of oriented gradients (COG) descriptor, which they further develop in their subsequent work by proposing LSS (latent support surfaces) \citep{ren_3d_2018} and COG~2.0 \citep{ren_clouds_2020}. Additionally, the approach was also adopted by \citet{liu_2-d_2018}. In general, the COG feature is able to describe complex 3D appearances within every orientation, as it consists of a gradient computation, 3D orientation bins and a normalization. For each proposal, the point cloud density, surface normal features and a COG-feature are calculated in a sliding window fashion, which are then scored with pre-trained SVMs.

In addition, manually configured features are typically used in combination with matching algorithms for detection purposes. For example, \citet{yamazaki_discovering_2018} create gradient-based image features by applying principal component analysis to the point cloud, which are subsequently used to compute the normalized cross-correlation. The main idea is to use the spatial relationships between image projection directions to discover the optimal template matching for detection. Similarly, \citet{teng_surface-based_2014} use handcrafted features, namely color histogram and key point histogram for template matching purposes. Another example is the approach proposed by \cite{he_3d_2017}. The authors create silhouette gradient orientations from RGB and surface normal orientations from depth images.

\subsection{Monocular Feature Learning}
\label{section:Feature Extraction_Deep Learning_Mono}
Probably the most difficult form of feature extraction for 3DOD is exercised in monocular representation. Since there is no direct depth data, well-informed features are difficult to obtain, and thus most approaches attempt to compensate for this lack of information with various depth substitution techniques.

\subsubsection{Solely Monocular Approaches}
\label{section:Feature Extraction_Deep Learning_Solely Mono}
In our literature corpus, there is only one paper which takes up the challenge of performing 3DOD exclusively with monocular images \citep{jorgensen_monocular_2019}. The authors use a single shot detection (SSD) framework (\cite{liu_ssd_2016}, see also Section \ref{section:Detection Module_single stage}) to generate per-object canonical 3D bounding box parameters. They start from a classical bounding box detector and add new output heads for 3DOD features. More specifically, they add distance, orientation, dimension and 3D corners to the already available class score and 2D bounding box heads. The conducted feature extraction is the same as it is in the 2D framework.

Models that use only monocular inputs follow the straightforward way of predicting spatial parameters directly without any range information. While the objectives of dimension and pose estimation are relatively easy to fulfill as they rely heavily on the given feature of appearance in 2D images, they face the challenge of difficult location estimation, since monocular inputs do not have a natural feature for spatial location \citep{leal-gupta_3d_2019, liu_deep_2019}.

\subsubsection{Informed Monocular Approaches}
\label{section:Feature Extraction_Deep Learning_Informed Mono}
The absence of depth information cannot be fully compensated for in purely monocular approaches. Therefore, many state-of-the-art monocular models supplement the 2D representation by an auxiliary depth estimation network or additional external data and information. The idea is to use prior knowledge of the target objects and the scene, such as shape, context and occlusion patterns to compensate for missing depth data.

These depth substitution techniques can be roughly divided into (i) \textit{depth estimation networks}, (ii) \textit{geometric constraints} and (iii) \textit{3D template matching}. Often, they are used in combination. 

While depth estimation networks are applied directly to the original representation to generate a new input for the model, geometric constraints and 3D model matching tackle the lack of range information in the later detection steps of the pipeline.

\paragraph{Depth Estimation Networks}
Depth estimation networks generate informed depth features or even entirely new representations that possess range information, such as point clouds derived synthetically from monocular imagery. These new representations are then subsequently exploited by depth-aware models. A representative model is Mono3D \citep{chen_monocular_2016}. It uses additional instance and semantic segmentation along with further features to reason about the pose and location of 3D objects.

\cite{srivastava_learning_2019} modify the generative adversarial network of BirdNet \citep{beltran_birdnet:_2018} to create a BEV projection from the monocular representation. All following operations such as feature extractions and predictions are then performed on this new representation.

Similarly, \citet{roddick_orthographic_2018} transform a monocular representation to the BEV perspective. They introduce an orthographic feature transformation network that maps the features from the RGB perspective to a 3D voxel map. The features of the voxel map are eventually reduced to the 2D orthographic BEV feature map by consolidation along the vertical dimension.

\cite{payen_de_la_garanderie_eliminating_2018} adapt the monocular depth recovery technique of \cite{godard_unsupervised_2017}, called Mono Depth, for their special case of monocular 360° panoramic processing. They predict a depth map by training a CNN on left-right consistency inside stereo image pairs. However, at the time of inference, the model only requires single monocular images to estimate a dense depth map.

Even more advanced, a few approaches devote themselves to generate a point cloud from monocular images. \cite{xu_multi-level_2018} use a stand-alone network for disparity prediction which is similarly based on Mono Depth to predict a depth map. Unlike \cite{payen_de_la_garanderie_eliminating_2018}, they further process the depth map to estimate a LiDAR-like point cloud.

An almost identical procedure is presented by \citet{ma_accurate_2019}. First, they generate a depth map through a self-defined CNN and then proceed to generate a point cloud by using the camera calibration files. While \citet{xu_multi-level_2018} mainly take the depth data as auxiliary information of RGB features, \citet{ma_accurate_2019} focus on taking the generated depth as a core feature and explicitly using its spatial information.

Similarly, \citet{weng_monocular_2019} adapt the deep ordinal regression network (DORN) by \citet{fu_deep_2018} for their pseudo LiDAR point cloud generation and then exploit a Frustum-PointNet-like model \citep{qi_frustum_2018} for the object detection task. Further information on the Frustum PointNet approach is given in Section \ref{section:Fusion_Cascaded Fusion}.

Instead of covering the entire scene in a point-based representation, \citet{ku_monocular_2019} primarily reduce the space by lightweight predictions and then only transform the candidate boxes to point clouds, preventing redundant computation. To do so, they exploit instance segmentation and available LiDAR data for training to reconstruct a point cloud in a canonical object coordinate system. A similar approach for instance depth estimation is pursued by \citet{qin_monogrnet_2019}.

\paragraph{Geometric Constraints}
Depth estimation networks offer the advantage of closing the gap of missing depth in a direct way. Yet, errors and noise occur during depth estimation, which may lead to biased overall results and contributes to a limited upper-bound of performance \citep{brazil_m3d-rpn_2019, barabanau_monocular_2020}.
Hence, various methods try to skip the naturally ill-posed depth estimation and tackle monocular 3DOD as a geometrical problem of mapping 2D into 3D space.

Especially in autonomous driving applications, the 3D box proposals are often constrained by a flat ground assumption, namely the street. It is assumed that all possible targets are located on this plane, since automotive vehicles do not fly. Therefore, these approaches force the bounding boxes to lay along the ground plane. In indoor scenarios, on the other hand, objects are located on various height levels. Hence, the ground plane constraint does not get the attention as in autonomous driving applications. Nevertheless, plane fitting is frequently applied in indoor scenarios to get the room orientation.

\citet{zia_towards_2015} were one of the first to assume a common ground plane in their approach, which helped them to extensively reconstruct the scene. Further, a ground plane drastically reduces the search space by only leaving two degrees of freedom for translation and one for rotation. 
Other representative examples for implementing ground plane assumptions in monocular representations are given by \citet{chen_monocular_2016}, \citet{du_general_2018} and \citet{leal-gupta_3d_2019}. All of them leverage the random sample consensus approach by \citet{fischler_random_1981}, a popular technique that is applied for ground plane estimation.

A different geometrical approach that is used to recover the under-constrained monocular 3DOD problem is to establish consistency between the 2D and 3D scenes. 
\citet{mousavian_3d_2017}, \citet{li_gs3d_2019}, \citet{liu_deep_fqnet_2019} and \citet{naiden_shift_2019} do so by projecting the 3D bounding box onto a previously determined 2D bounding box. The core notion is that the 3D bounding box should fit tightly to at least one side of its corresponding 2D box detection. \cite{naiden_shift_2019} use, for instance, a least square method for the fitting task.

Other methods deploy 2D-3D consistency by incorporating geometric constraints such as room layout and camera pose estimations through an entangled 2D-3D loss function \cite[e.g.,][]{huang_cooperative_2018,simonelli_disentangling_2019,brazil_m3d-rpn_2019}. For example, \citet{huang_cooperative_2018} define the 3D object center through corresponding 2D and camera parameters. Then a physical overlap between 3D objects and 3D room layout gets penalized. \cite{simonelli_disentangling_2019} first disentangle the 2D and 3D detection losses to optimize each loss individually. Subsequently, they also leverage the correlation between 2D and 3D in a combined multi-task loss.

Another approach is presented by \citet{qin_triangulation_2019}. The authors exploit triangulation, which is well known for estimating 3D geometry in stereo images. They use 2D detection of the same object in a left and right monocular image for a newly introduced anchor triangulation, where they directly localize 3D anchors based on the 2D region proposals.

\paragraph{3D template matching}
An additional way of handling monocular representations for 3DOD is to match the images with 3D object templates. The idea is to have a database of object images from different viewpoints and their underlying 3D depth features. One popular approach for creating templates is to render synthetic images from computer-aided design (CAD) models, whereby images are created from all sides of the object. Then the monocular input image is searched and matched using this template database. On this basis, the object pose and location can be concluded. 

\citet{fidler_3d_2012} address the task of monocular object detection with the representation of an object as a deformable 3D cuboid. The 3D cuboid consists of faces and parts, which are allowed to deform according to their anchors in the 3D box. Each of these faces is modelled by a 2D template that corresponds to appearance of the object from an orthogonal point of view. It is assumed that the 3D cuboid can be rotated so that the image view from a defined set of angles can be projected onto the respective cuboids' face and subsequently scored by a latent SVM.

\cite{chabot_deep_2017} initially use a network to generate 2D detection results, vehicle part coordinates and a 3D box dimension proposal. Thereafter, they match the dimensions to a 3D CAD dataset consisting of a fixed number of object models to assign the corresponding 3D shapes in the form of manually annotated vertices. Those 3D shapes are then used for performing 2D-to-3D pose matching in order to recover 3D orientation and location. 
\cite{barabanau_monocular_2020} reason their approach on sparse but salient features, namely 2D key points. They match CAD templates based on 14 key points. After assigning one of five distinct geometric classes, they execute an instance depth estimation by a vertical plane passing through two of the visible key points to lift the predictions into a 3D space. 

3D templates provide a potentially powerful source of information. However, a sufficient number of models are not always available for each object class. Therefore, these methods tend to focus on a small amount of classes. The limitation to shapes covered by the selection of 3D templates makes it difficult to generalize or extend 3D template matching to classes for which no models are available \citep{ku_monocular_2019}.

In summary, monocular 3DOD achieves promising results. Nevertheless, the lack of depth data prevents monocular 3DOD from reaching state-of-the-art results. The depth substitution techniques may limit the detection performance because errors in depth estimation, geometrical assumptions or template matching are propagated to the final 3D box prediction.
In addition, informed monouclar approaches may be comparably vulnerable to external attacks. {\citet{Cheng_zhiyuan_2022}} shows that both physical and digital attacks on depth estimation networks have serious impacts on 3DOD performance.

\subsection{Point-wise Feature Learning}
\label{section:Feature Extraction_Pointwise}
As described above, the application of deep learning techniques on point cloud data is not straightforward due to the irregularity of the point cloud (cf. Section \ref{section:Input_Pointcloud}). Many existing methods try to leverage the expertise of convolutional feature extraction by either projecting point clouds onto 2D image views or by converting them into regular grids of voxels. However, projecting point clounds onto a specific viewpoint discards valuable information, which is particularly important in crowded scenes. The voxelization, on the other hand, leads to high computational costs due to the sparse nature of point clouds and also suffers from information loss in point-crowded voxels. Either way, manipulating the original data may have a negative effect.

To overcome the problem of irregularity in point clouds with an alternative approach, \citet{qi_pointnet_2017} proposed \textit{PointNet}, which is able to learn point-wise features directly from the raw point cloud. It is based on the assumption that points which lie close to each other can be grouped together and compressed as a single point. Shortly after, \citet{qi_pointnet++_2017-1} introduced its successor \textit{PointNet++}, adding the ability to capture local structures in the point cloud. Both networks are originally designed for classification tasks on the whole point cloud, next to being able of predicting semantic classes for each point of the point cloud. Thereafter, \cite{qi_frustum_2018} introduced a way to implement PointNet into 3DOD by proposing Frustum-PointNet. By now, many of the state-of-the-art 3DOD methods are based on the general PointNet-architectures. Therefore, it is crucial to understand the underlying architecture and how it is used in 3DOD methods.

\subsubsection{PointNet}
\label{section:Feature Extraction_PointNet}

PointNet consists of three key modules: (i) a max-pooling layer serving as a symmetric function, (ii) a local and global information combination structure in the form of a multi-layer perceptron (MLP) and (iii) two joint alignment networks for the alignment of input points and point features, respectively \citep{qi_pointnet_2017}.

To deal with the permutation invariance of point clouds, PointNet is built with symmetric functions in the form of max-pooling operations. Symmetric functions have the same output regardless of the input order. The max-pooling operation results in a global feature vector that aggregates information from all points of the point cloud. Since the max pooling function operates as a “winner takes it all” paradigm, it does not consider local structures, which is the main limitation of PointNet.

Following, PointNet accommodates an MLP that uses the global feature vector subsequently for the classification tasks. Other than that, the global features can also be used in a combination with local point features for segmentation purposes.

The joint alignment networks ensure that a single point cloud is invariant to geometric transformations (e.g., rotation and translation). PointNet uses this natural solution to align the input points of a set of point clouds to a canonical space by pose normalization through spatial transformers, called T-Net. The same operation is deployed again by a separate network for feature alignment of all point clouds in a feature space. Both operations are crucial for the networks' predictions to be invariant of the input point cloud.

\subsubsection{PointNet++}
\label{section:Feature Extraction_PointNet++}

\begin{figure*}[htbp]
\centering
\includegraphics[width=0.85\textwidth]{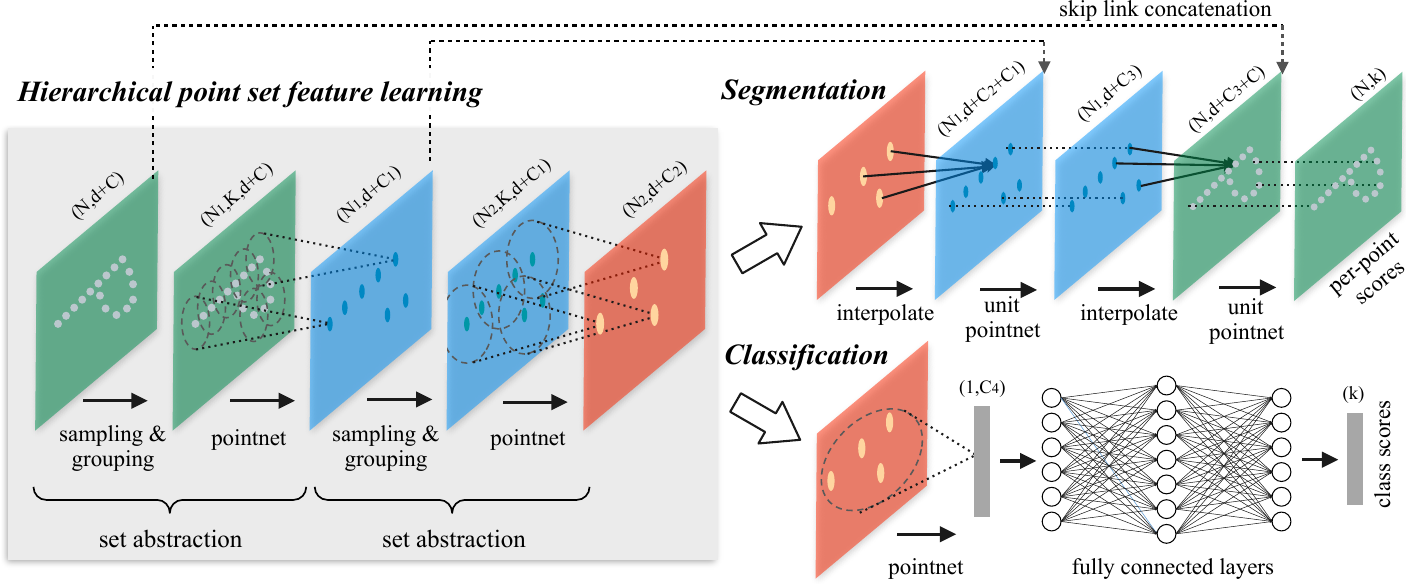}
\caption[Architecture of PointNet++] {Architecture of PointNet++ \citep{qi_pointnet++_2017-1}\label{fig:PointNet++}}
\end{figure*}

PointNet++ is built in a hierarchical manner on several set abstraction layers to address the original PointNet's missing ability to consider local structures. At each level, a set of points is further abstracted to produce a new set with fewer elements, in fact summarizing the local context. The set abstraction layers are composed again of three key layers: (i) a sampling layer, (ii) a grouping layer, (iii) and a PointNet layer \citep{qi_pointnet++_2017-1}. Figure~\ref{fig:PointNet++} provides an overview of the architecture.

The sampling layer is employed to reduce the resolution of points. PointNet++ uses farthest point sampling (FPS) which only samples the points that are the most distant from the rest of the sampled points \citep{bello_review_2020}. Thereby, FPS identifies and retains the centroids of the local regions for a set of points.

Subsequently, the grouping layers are used to group the representative points, which are obtained from the sampling operation, into local patches. Moreover, it constructs local region sets by finding neighboring points around the sampled centroids. These are further exploited to compute the local feature representation of the neighborhood. PointNet++ adopts ball query, which searches a fixed sphere around the centroid point and then groups all within laying points.

The grouping and sampling layers represent preprocessing tasks to capture local structures before the abstracted points are passed to the PointNet layer. This layer consists of the original PointNet-architecture and is applied to generate a feature vector of the local region pattern. The input for the PointNet layer is the abstraction of the local regions, i.e. the centroids and local features that encode the centroids' neighborhood. 

The process of grouping, sampling and applying PointNet is repeated in a hierarchical fashion, with points being down-sampled further and further until the last layer yieals a final global feature vector \citep{bello_review_2020}. In this way, PointNet++ can work with the same input data at different scales and generate higher level features at each set abstraction layer, while thus capturing local structures. 

\subsubsection{PointNet-based Feature Extraction}
\label{section:Pointwise_using_PointNet}

In the following, no explicit distinction is made between PointNet and PointNet++. Instead, we summarize both under the term PointNet-based approaches, considering that we primarily want to imply that point-wise methods for feature extraction or classification purposes are used. 

Many state-of-the-art models use PointNet-like networks in their pipeline for feature extraction. An exemplary selection of seminal works include the proposals from \citet{qi_frustum_2018},  \citet{yang_ipod_2018}, \citet{zhou_voxelnet_2018} \citet{xu_pointfusion:_2018}, \citet{shi_pointrcnn_2019}, \citet{shin_roarnet_2019}, \citet{pamplona_pointnet_2019}, \citet{lang_pointpillars_2019}, \citet{wang_frustum_2019-1}, \citet{yang_3dssd_2020}, \citet{li_3d_2020}, \citet{yoo_3d-cvf_2020}, \citet{zhou_joint_2020-1} and \citet{huang_epnet_2020}. 

Very little is changed in the way PointNet is used from the models adopting it for feature extraction, indicating that it is already a well-designed and mature technique. Yet, \citet{yang_3dssd_2020} examine in their work that the up-sampling operation in the feature propagation layers and refinement modules consume about half of the inference time of existing PointNet approaches. Therefore, they abandon both processes to drastically reduce inference time. However, predicting only on the surviving representative points on the last set abstraction layer leads to huge performance drops. Therefore, they propose a novel sampling strategy based on feature distance and merge this criterion with the common Euclidean distance sampling for meaningful features. 

PointNet-based 3DOD models generally show superior performances in classification as compared to models using other feature extraction methods. The set abstraction operation brings the crucial advantage of flexible receptive fields for feature learning through setting different search radii within the grouping layer \citep{shi_pointrcnn_2019}. Flexible receptive fields can better capture the relevant content or features because they adapt to the input. Fixed receptive fields such as convolutional kernels are always limited in their dimensionality, which can mean that features and objects of different sizes cannot be captured so well.. However, PointNet operations, especially set abstractions, are computationally expensive, which translates into long inference times compared to convolutions or fully connected layers \citep{yang_std_2019, yang_3dssd_2020}.

\subsection{Segment-wise Feature Learning}
\label{section:Feature Extraction_Segmentwise}
Segment-wise feature learning follows the idea of a regularized 3D data representation. In comparison to point-wise feature extraction, it does not take the points of the whole point cloud into account as in the previous section. Instead it processes an aggregated set of grid-like representations of the 3D scene (cf. Section~\ref{section:Data Representation_Gridcells}), which are therefore considered as \textit{segments}.

In the following, we describe central aspects and operations of segment-wise feature learning, including (i)~\textit{feature initialization} (Section~\ref{section:Feature Extraction_Feature Initalization}), (ii)~\textit{2D and 3D convolutions} (Section~\ref{section:Feature Extraction_2D-3D-Conv}), \textit{sparse convolution} (Section~\ref{section:Feature Extraction_Sparse Conv}) and \textit{voting scheme} (Section~\ref{section:Feature Extraction_Voting}).

\subsubsection{Feature Initialization}
\label{section:Feature Extraction_Feature Initalization}

Volumetric approaches discretize the point cloud into a specific volumetric grid during preprocessing. Projection models, on the other hand, typically lay a relatively fine-grained grid on the 2D mapping of the point cloud scene. In either case, the representation is transformed into segments. In other words, segments can be either be volumetric grids like voxels and pillars (cf. Section~\ref{section:Data Representation_volume_based}) or discretized projections of the point cloud like RV and BEV projections (cf. Section \ref{section:Data Representation_projection}). 

These generated segments enclose a certain set of points that does not yet have a processable state. Therefore, an encoding is applied on the individual segments that aggregates the points they enclose.

The intention of the encoding is to fill the formulated segments or grids with discriminative features that provide information about the set of points that lie in each individual grid. This process is called feature initialization. Through the grid, these features are now available in a regular and structured format. In contrast to handcrafted feature extraction methods (cf. Section~\ref{section:Feature Extraction_Handcrafted}), these grids are not yet used for detection, but are made accessible to CNNs or other extraction mechanisms to further condense the features.

For volumetric approaches, current research can be divided into two streams of feature initialization. The first and probably more intuitive approach is to manually encode the voxels. However, handcrafted feature encoding introduces a bottleneck that discards spatial information and may prevent these approaches from effectively utilizing 3D shape information. This led to the latter approach, where models apply a lightweight PointNet to each voxel to learn point-wise features and assign them in aggregated form as voxel features, which is referred to as \textit{voxel feature encoding} (VFE) \citep{zhou_voxelnet_2018}.

\paragraph{Volumetric Feature Initialization}
\label{section:Feature Extraction_Volumetric}
Traditionally, voxels are encoded into manually selected features such that each voxel contains one or more values consisting of statistics computed from the points within that voxel cell. The selection of features is already a crucial task, as they should capture the important information within a voxel and describe the corresponding points in a sufficiently discriminative way.

Early approaches such as Sliding Shapes \citep{song_sliding_2014} use a combination of four types of 3D features to encode the cells, namely point density, a 3D shape feature, a surface normal feature and a specifically designed feature to deal with the problem of self-occlusion, called the truncated signed distance function. Others, however, rely only on statistical encoding like \citet{wang_voting_2015} as well as their adaption by \citet{engelcke_vote3deep_2017}, which propose three shape factors, the mean and variance of the reflectance values of points as well as a binary occupancy feature. 

Much simpler approaches are pursued by \citet{li_3d_2017} and \citet{li_3d_guivant_2019} using a binary encoding to express whether a voxel contains points or not. To avoid too much information loss due to a rudimentary binary encoding, the voxel size is usually chosen comparatively small to generate high-resolution 3D grids \citep{li_3d_guivant_2019}.

More recently, voxel feature initialization has shifted more and more towards deep learning approaches for similar reasons as feature extraction in other computer vision tasks, where manually selected features are not as performant as learned ones.

While segment-wise approaches prove to be comparatively efficient 3D feature extraction methods, point-wise models show impressive results in detection accuracy as they have recourse to the full information of each point of the point cloud. By manually encoding the voxels into standard features, a lot of information is usually lost.

As a response, \citet{zhou_voxelnet_2018} introduced the seminal idea of VFE and a corresponding deep neural network which moves the feature initialization from hand-crafted voxel feature encoding to deep-learning-based encoding. More specifically, they proposed VoxelNet, which is able to extract point-wise features from each segment of a voxelized point cloud through a lightweight PointNet-like network. Subsequently, the individual point features are stacked together with a locally aggregated feature at voxel level. Finally, this volumetric representation is fed into 3D convolutional layers for further feature aggregation. 

\begin{figure}[htbp]
\centering
\includegraphics[width=1\columnwidth]{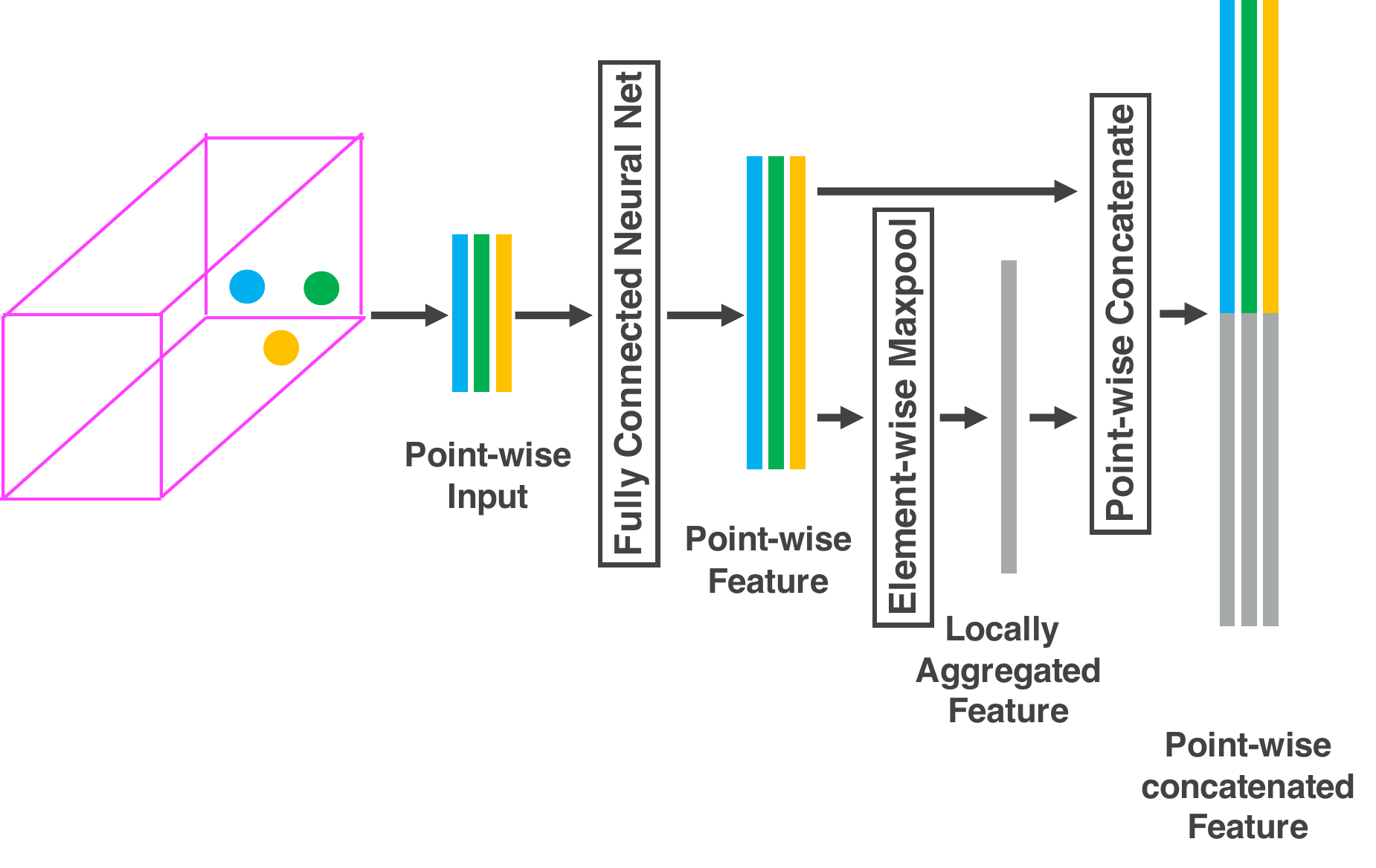}
\caption[Architecture of Voxel Feature Encoding Module] {Architecture of VFE module in VoxelNet \citep{zhou_voxelnet_2018}\label{fig:VFE}}
\end{figure}

The use of PointNet allows VoxelNet to capture inter-point relations within a voxel and therefore hold a more discriminative feature then normal encoded voxel consisting of statistical values. A schematic overview of VoxelNet's architecture is shown in Figure \ref{fig:VFE}.

The seminal idea of VFE is used – in modified versions – in many subsequent approaches. Several works focus on improving performance and efficiency of VoxelNet. In terms of performance, \citet{kuang_voxel-fpn_2020} developed a novel feature pyramid extraction paradigm. To speed up the model, \citet{yan_second_2018} and \citet{shi_points_2020} combined VFE with more efficient sparse convolutions. In addition, \citet{sun_3d_2018} and \citet{chen_fast_2019} reduced the original architecture of VFE to reduce inference times.

\paragraph{Projection-based Feature Initialization}
\label{section:Feature Extraction_Projection}

In projection approaches, feature initialization of cells is usually done by hand. Both RV and BEV utilize fine-grained grids that are primarily filled with statistical quantities of the point lying within them. Only recently have the first models begun to encode features using deep learning methods \cite[e.g.,][]{lehner_patch_2019,wang_multi-view_2020,liang_rangercnn_2020}.

For RV, it is most popular to encode the projection map into three-channel features, namely height, distance and intensity \citep{chen_multi-view_2017,zhou_fvnet_2019,liang_rangercnn_2020}. 
Instead, \citet{meyer_lasernet_2019} and \citet{meyer_sensor_2019} form a five-channel image with range, height, azimuth angle, intensity and a flag indicating whether a cell contains a point. 
In contrast to manual encoding, \citet{wang_multi-view_2020} use a point-based, fully connected layer to learn high-dimensional point features of the LiDAR point cloud, and then apply a max-pooling operation along the $z$-axis to obtain the cells' features in RV.

Similar to RV, \citet{chen_multi-view_2017} encode each cell of a BEV representation in height, intensity and density. To increase significance of this representation, the point cloud is divided into $M$ slices along the $y$-axis, resulting in a BEV map with $M + 2$ channel features.

After the introduction to 3DOD by \cite{chen_multi-view_2017}, BEV representation became quite popular and many other approaches followed their proposal of feature initialization \cite[e.g.,][]{beltran_birdnet:_2018,liang_deep_2018,wang_fusing_2018,simon_complex-yolo_2019, li_one-stage_2019_hu}. Yet, others choose a simpler set up, encoding only the maximum height and density without slicing the point cloud \citep{ali_yolo3d_2019} or even using a binary occupancy encoding \citep{yang_pixor_2018, he_structure_2020}.

To avoid information loss, more recent approaches first extract features using deep learning approaches and then project these features into BEV \cite[e.g.,][]{wang_multi-view_2020, liang_rangercnn_2020}.

Analogous to their RV approach, \citet{wang_multi-view_2020} again use the learned point-wise features of the point cloud, but now apply the max pooling operation along the $y$-axis for feature aggregation in BEV. \cite{liang_rangercnn_2020}, on the other hand, first extract features in RV and then transform them into a BEV representation, adding more high-level information compared to directly projecting the point cloud to BEV. 

After feature initialization of either volumetric grids or projected cells, segment-wise solutions usually utilize 2D/3D convolutions to extract features of the global scene. 

\subsubsection{2D and 3D Convolutions}
\label{section:Feature Extraction_2D-3D-Conv}

%\paragraph{2D Convolutions}
Preprocessed 2D representations such as feature-initialized projections, monocular images and RGB-D (2D) images all have the advantage that they can all leverage mature 2D convolutional techniques to extract features.

%\paragraph{3D Convolutions}
Volumetric voxel-wise representations, on the other hand, represent the spatial space in a regular format which are accessed by 3D convolutions. However, directly applying convolutions in 3D space is a very inefficient procedure due to the multiplication of space. 

Early approaches to extend the traditional 2D convolutions to 3D were applied by \citet{song_deep_2016} as well as \citet{li_3d_2017} by placing 3D convolutional filters in 3D space and performing feature extraction in an exhaustive operation. Since the search space increases drastically from 2D to 3D, this procedure involves immense computational costs.

Further examples using conventional 3D CNNs can be found in the models of \citet{chen_3d_2015}, \citet{sun_3d_2018}, \citet{zhou_voxelnet_2018} and \citet{sindagi_mvx-net_2019}. 

Despite delivering state-of-the-art results, the conventional 3D CNN lacks efficiency. Given the fact that the sparsity of point clouds leads to many empty and non-discriminative voxels in a volumetric representation, the exhausting 3D CNN operations perform a large amount of redundant computations. This issue can be addressed by a sparse convolution.

\subsubsection{Sparse Convolution}
\label{section:Feature Extraction_Sparse Conv}

Sparse convolution was proposed by \cite{graham_spatially-sparse_2014} and \cite{graham_sparse_2015}. First, a ground state is defined for the input data. The ground state expresses whether the spatial location (site) is active or not. A site in the input representation is active if it has a non-zero value, which in the case of a regularized point cloud is a voxel enclosing at least a certain threshold of points. Furthermore, a site in the following layers is active if any of the spatial locations from the foregoing layer, from which it receives its input, is active. Therefore, sparse convolution must only process the sites which differ from the ground state of the preceding convolution layer, focusing computational power on the meaningful and new information.

To reduce resource costs and speed up feature extraction, irrelevant regions need to be skipped when processing point clouds. \cite{yan_second_2018} were the first to apply sparse convolutions in 3DOD, which do not suffer but exploit sparsity.

However, the principle of sparse convolution has the disadvantage that it continuously leads to dilation of the data, as it discards any non-active sites. The deeper a network becomes, the more the sparsity is reduced and the data is dilated. For this reason, \citet{graham_3d_2018} introduced submanifold sparse convolution, where the input is first padded so that the output retains the same dimensions. Moreover, the output is active only when the central site of the receptive field is active, thus preserving the efficiency of sparse convolution while simultaneously maintaining sparsity.

Relevant works using sparse 3D convolutions are proposed by \citet{yan_second_2018}, \citet{chen_fast_2019}, \citet{shi_points_2020}, \citet{pang_clocs_2020}, \citet{yoo_3d-cvf_2020}, \citet{he_structure_2020}, \citet{zheng_cia-ssd_2020} and \citet{deng_voxel_2021}.

\subsubsection{Voting Scheme}
\label{section:Feature Extraction_Voting}
Another approach to exploit sparsity of point clouds is to apply a voting scheme as exemplified by \citet{wang_voting_2015}, \citet{engelcke_vote3deep_2017} and \citet{qi_deep_2019}. The idea behind voting is to let each non-zero site in the input declare a set of votes to its surrounding cells in the output layer. The voting weights are computed by flipping the convolutional weights of the filter along its diagonal \citep{fernandes_point-cloud_2021}. As with sparse convolution, processing only needs to be performed on non-zero sites. Hence, computational cost is proportional to the number of occupied sites rather than to the dimension of the scene. 
Facing the same problem of dilation as sparse convolution, \citet{engelcke_vote3deep_2017} argue to select non-linear activation functions. In this case, rectified linear units help to maintain sparsity because only features with values greater than zero, and not just non-zero, are allowed to cast votes. 

Mathematically, the feature-centric voting is equivalent to the submanifold sparse convolution, as \citet{wang_voting_2015} proof in their work.

\section{Fusion Approaches}
\label{section:Fusion}

Single modality pipelines for 3DOD have developed well in recent years and have shown remarkable results. Yet, unimodal models still reveal shortcomings preventing them to reach full maturity and human-like performance. For instance, camera images lack depth information and suffer from truncation and occlusion, while point clouds lack texture information and are sparse at longer distances. To overcome these problems, recent research is increasingly focusing attention on fusion models that attempt to leverage the combination of information from different modalities.

The main challenges of fusion approaches are the synchronization of the different representations and the preservation of relevant information during the fusion process. Further, holding the additional complexity at a computationally reasonable level must also be taken into account. 

Fusion methods can be divided into two classes depending on the orchestration of the modality integration, namely (i)~\textit{cascaded fusion} (Section~\ref{section:Fusion_Cascaded Fusion}) and (ii)~\textit{feature fusion} (Section~\ref{section:Fusion_Feature Fusion}). 
The former combines different sensor data and their individual features or predictions across different stages, whereas the latter jointly reasons about multi-representation inputs.

\subsection{Cascaded Fusion}
\label{section:Fusion_Cascaded Fusion}

\begin{figure*}[htbp]
\centering
\includegraphics[width=0.8\textwidth]{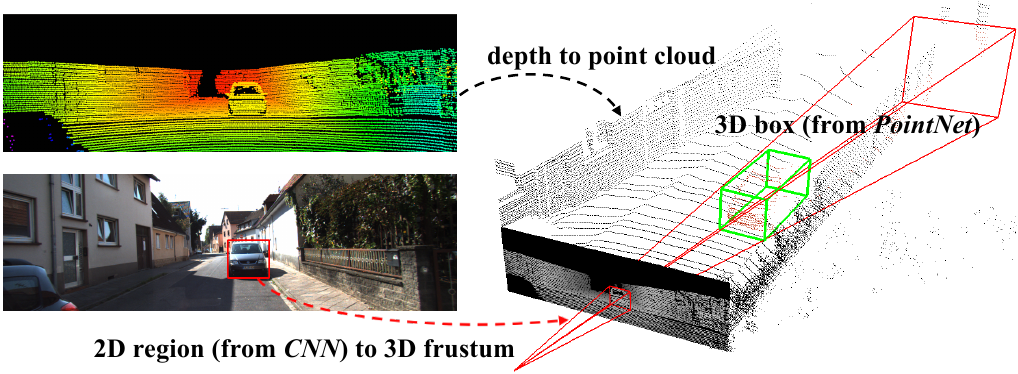}
\caption[Cascaded Fusion Scheme of Frustum-PointNets] {Cascaded fusion scheme of Frustum-PointNets \citep{qi_frustum_2018}\label{fig:Cascaded Fusion}}
\end{figure*}

Cascaded fusion methods use consecutive single-modality detectors to restrict second-stage detection by the results of the first detector. Typically, monocular-based object detectors are leveraged in the first stage to define a contained subset of the point cloud containing only 3D points that are likely to define an object. Hence, in the second stage, 3D detectors only need to reason over a limited 3D search space.

Two seminal works in this regard are the fusion frameworks proposed by \cite{lahoud_2d-driven_2017} and \cite{qi_frustum_2018}. The approaches use the detection results from the 2D image to extrude a corresponding frustum into 3D space. For each 2D proposal, a frustum is generated. The popular Frustum-PointNets by \cite{qi_frustum_2018} then processes the frustum with PointNet for instance segmentation. Finally, the amodal 3D box is predicted based on the frustum and the extracted foreground points (see Figure \ref{fig:Cascaded Fusion}). \cite{lahoud_2d-driven_2017}, on the other hand, first estimates the orientation for each object within the frustum. In the last step, they apply an MLP regressor for the 3D boundaries of the object. 

Several approaches follow this basic idea in a similar way. For example, both \citet{yang_ipod_2018} and \citet{ferguson_2d-3d_2019} use 2D semantic segmentation and then project these foreground pixels of the image into a point cloud. The selected points are subsequently exploited for proposal generation through PointNet or convolution operations. \citet{du_general_2018} leverage the restricted 3D space by applying a model matching algorithm for detection purpose. In contrast, \citet{shin_roarnet_2019} attempt to improve the 3D subset generation by creating point cloud region proposals with the shape of standing cylinders instead of frustums, which is more robust to sensor synchronization. 

While the described models above mainly focus on the frustum creation process, \citet{wang_frustum_2019-1}, \citet{zhang_faraway-frustum_2020} as well as \citet{shen_frustum_2020} seek to advance the processing of the frustums.

Due to its modular nature, Frustum-PointNet is not able to provide an end-to-end prediction. To overcome this limitation, \citet{wang_frustum_2019-1} subdivide the frustums to eventually use of a fully convolutional network allowing a continuous estimation of oriented boxes in 3D space. They generate a sequence of frustums by sliding along the frustum axis, and then aggregate the grouped points of each respective section into local point-wise features. These features on frustum-level are arranged as a 2D feature map, enabling the use of a subsequent fully convolutional network.

Other than that, \citet{shen_frustum_2020} aim to integrate the advancements of voxelization into frustum approaches by transforming regions of interests (ROIs) within the point frustums into 3D volumetric grids. Thus, only relevant regions are voxelized, allowing a high resolution that improves the representation while still being efficient. In this case, the voxels are then fed to a 3D fully convolutional network.

More recently, \citet{zhang_faraway-frustum_2020} observed that point-based 3DOD does not perform well in longer range because of an increasing sparsity of point clouds. Therefore, they take advantage of RGB images that contain enough information to recognize distant objects with mature 2D detectors. While following the idea of frustum generation, the estimated location of objects that are considered to be distant are recognized. Taking into account that very few points define these objects in a point cloud, they do not possess sufficient discriminative information for neural networks, so that 2D detectors are applied on corresponding images. Otherwise, for close objects, \citet{zhang_faraway-frustum_2020} use conventional neural networks to process the frustum.

\subsection{Feature Fusion}
\label{section:Fusion_Feature Fusion}

Since the performance of cascaded fusion models is always limited by the accuracy of the detector at each stage, some researchers try to increase performance by arguing that the models should infer more jointly across modalities.

\begin{figure}[htbp]
\centering
\includegraphics[width=1\columnwidth]{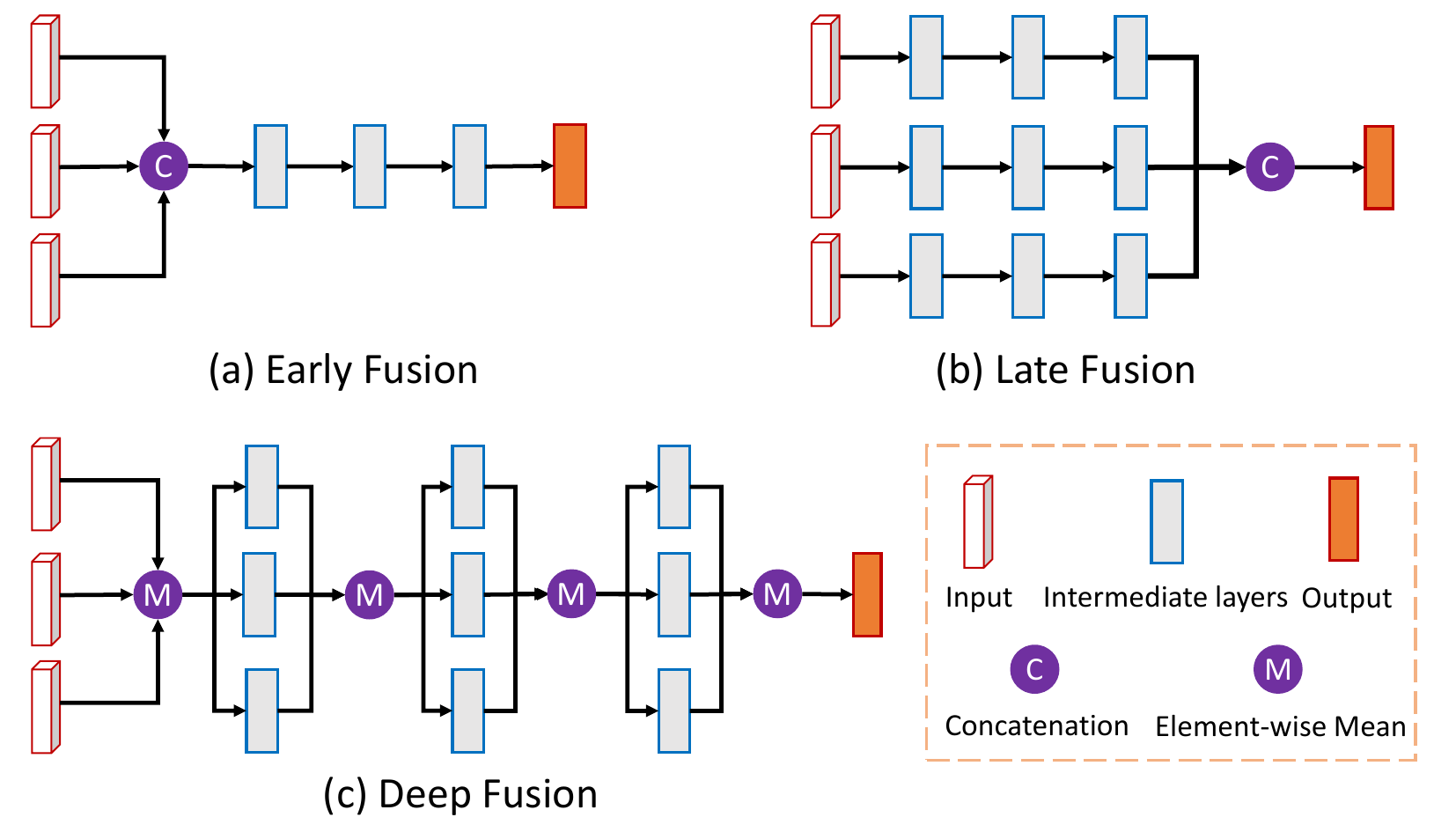}
\caption[Feature Fusion Schemes] {Early, late and deep feature fusion scheme \citep{chen_multi-view_2017}\label{fig:Feature Fusion}}
\end{figure}

To this end, feature fusion methods first concatenate the information from different modalities before reasoning about the combined features, trying to exploit the diverse information in their combination. Within feature fusion, it can be further distinguished between (i) early, (ii) late and (iii) deep fusion approaches \citep{chen_multi-view_2017}, constituting at which stage of the 3DOD pipeline fusion occurs. Figure~\ref{fig:Feature Fusion} provides an illustrative overview of the different fusion schemes.

\textit{Early fusion} merges multi-view features in the input stage before any feature transformation takes place, and proceeds with a single network to predict the results. \textit{Late fusion}, in contrast, uses multiple subnetworks that process the individual inputs separately up until the last stage of the pipeline, where they get concatenated in the prediction stage. Beyond that, \textit{deep fusion} allows an interaction of different input modalities at several stages in the architecture and alternately performs feature transformation and feature fusion.

Although the following approaches can all be categorized as feature fusion methods, the classification between the various subclasses of early, late and deep fusion is not trivial and can be fluid. Nevertheless, the concepts help to convey a better understanding of feature fusion processes.

The pioneers among 3DOD fusion approaches are \citet{chen_multi-view_2017}, introducing the multi-view approach. They take multiple perspectives, specifically RV, BEV and FV as input representations. The BEV representation is used to generate 3D candidates, followed by a region-wise feature extraction by projecting these 3D proposals onto the respective feature maps of each view. A deep fusion scheme is then used to combine the information element-wise over several intermediate stages.

\citet{ku_joint_2018} use 3D anchors mapped to both the BEV representation and the 2D image. Subsequently, a crop and resize operation is applied to every projected anchor, and the feature crops from both views are fused via an element-wise mean operation at an intermediate convolutional layer. Unlike \citet{chen_multi-view_2017}, \citet{ku_joint_2018} not only merge features in the late refinement stage but already in the region proposal stage to generate positive proposals. They specifically use the full resolution feature map to improve prediction quality, particularly for small objects. Similar approaches to \citet{chen_multi-view_2017} and \citet{ku_joint_2018} are performed by \citet{chen_3d_2018}, \citet{li_one-stage_2019_hu} and \citet{wang_multi-channel_2019}, who also fuse the region of interests of the input data representations element-wise. Yet, \citet{chen_3d_2018} use segment-wise 3D detection for box proposals in the first stage.

In contrast, \citet{rahman_3d_2019} not only fuse the regions of interest, but combine the entire feature map of processed monocular images and FV-representations already at the region proposal stage.

Further, \cite{ren_context-assisted_2018} want to leverage not only object detection information but also context information. Therefore, they simply concatenate the processed features of 2D scene classification, 2D object detection and 3D object detection of the voxelized scene before feeding them to a conditional random field model for joint optimization.

Rather than just combining regular representations through multi-view fusion models, some approaches also aim to merge raw point features with other representations.

For example, \citet{xu_pointfusion:_2018} process the raw point cloud with PointNet. They then concatenate each point-wise feature with a global scene feature and the corresponding image feature. Each point is then used as a spatial anchor to predict the offset to the 3D bounding box. Likewise, \citet{yang_std_2019} use features from a a PointNet++ backbone to extract semantic context features for each point. The sparse features subsequently get condensed by a point pooling layer to take advantage of a voxel-wise representation applying VFE.

\citet{shi_pv-rcnn_2020} first use 3D convolution on a voxel representation to summarize the scene into a small set of key points. In the next step, these voxel-feature key points are fused with the grid-based proposals for refinement purposes.

The previously presented models all perform late or deep fusion procedures. Instead of fusing multi-sensor features per object after the proposal stage, \citet{wang_fusing_2018} follow the idea of early fusing BEV and image views using sparse, non-homogeneous pooling layers over the full resolution. Similarly, \citet{meyer_sensor_2019} also employ early fusion, but they use RV images for the point-cloud-related representation. The approach associates the LiDAR point cloud with camera pixels by projecting the 3D points onto the 2D image. The image features are then concatenated and further processed by a fully connected convolutional network.

Furthermore, several works apply a hybrid approach of early and late fusion schemes. For example, \citet{sindagi_mvx-net_2019} first project LiDAR onto an RV representation and concatenate image features with the corresponding points in an early fusion fashion. Then they apply a VFE layer to the voxelized point cloud and append the corresponding image features for each non-empty voxel.
While early concatenation already fuses features, late fusion aggregates image information for volumetric-based representation, where voxels may contain low-quality information due to low point cloud resolution or distant objects.

Akin, \citet{liang_multi-task_2019} initially conduct feature fusion to RGB images and BEV feature maps. Thereby, they incorporate multi-scale image features to augment the BEV representation. In the refinement stage, the model fuses image and augmented BEV again, but other than in the first stage, the fusion occurs element-wise for the regions of interest. They further add ground estimation and depth estimation to the fusion framework to advance the fusion process. 

\citet{simon_complexer-yolo_2019} extend their previous work Complex YOLO \citep{simon_complex-yolo_2019} with Complexer You Only Look Once (YOLO) by exchanging the input of a BEV map for a voxel representation. To leverage all inputs, they first create a semantic segmentation of the RGB image and then fuse this picture point-wise on the LiDAR-frame to generate a semantic voxel grid.

Lately, \citet{zheng_cia-ssd_2020} first initialize features of a voxelized point cloud by calculating the mean coordinates and intensities of points in each voxel. They then apply sparse convolution and transform the representation into a dense feature map before condensing it on the ground plane to produce a BEV feature map.

\paragraph{Continuous Fusion}
LiDAR points are continuous and sparse, whereas cameras capture dense features at a discrete state. Fusing these modalities is not a trivial task due to the one-to-many projection. In other words, there is not a corresponding LiDAR point for every image pixel in every projection and vice versa.

To overcome this discontinuous mapping of images into point-cloud-based representations such as BEV, \citet{liang_deep_2018} propose a novel continuous convolution that is applied to create a dense feature map through interpolation. They propose to project the image feature map onto a BEV space and then fuse the original LiDAR BEV map in a deep fusion manner through continuous convolution over multiple resolutions. The fused feature map is then further processed by a 2D CNN to solve the discrepancy between image and projection representations.

\paragraph{Attention Mechanism}
A common challenge among fusion approaches is the occurrence of noise and the propagation of irrelevant features. Previous approaches simply fuse multiple features by concatenations or element-wise summation and/or mean operations. Thereby, noise such as truncation and occlusion gets inherited to the resulting feature maps. Thus, inferior point features will be obtained through fusion. Attention mechanisms can cope with these difficulties by determining the relevance of each feature to only fuse features that improve the representation.

\citet{lu_scanet_2019} use a deep fusion approach for BEV and RGB images in an element-wise way, but additionally incorporate attention modules over both modalities to leverage the most relevant features. Spatial attention adapts pooling operations over different feature map scales, while the channel-wise fusion applies global pooling. Both create an attention map that expresses the importance of each feature. These attention maps are then multiplied with the feature map and finally fused.

In analogy, \citet{wang_multi-view_2020} use an attentive point-wise fusion module to estimate the channel-wise importance of BEV, RV and image features. In contrast, they deploy the attention mechanism after the concatenation of the multi-view feature maps to consider the mutual interference and the importance of the respective features. They specifically address the issue of ill-posed information introduced by the front view of images and RV. To compensate the inevitable loss of geometric information through the projection of LiDAR points, the authors finally enrich the fused point features with raw point features through an MLP network. 

Consecutive to the attention fusion, \citet{yoo_3d-cvf_2020} use first stage proposals from the attentive camera-LiDAR feature map to extract the single modality LiDAR and camera features of the proposal regions. Using a PointNet encoding, these are subsequently fused element-wise with the joint feature map for refinement.

In contrast, \citet{huang_epnet_2020} operate directly on the LiDAR point cloud introducing a point-wise fusion. In their deep fusion approach, they process a PointNet-like geometric - and a convolution-based - image stream in parallel. Between each abstraction stage, the point features are fused with the semantic image features of the corresponding convolutional layer by applying an attention mechanism.

Moreover, \citet{pang_clocs_2020} observed that element-wise fusion takes place after non-maximum suppression (NMS), which can which can result in useful candidates of each modality being incorrectly suppressed. NMS is used to suppress duplicate candidates after proposal and prediction stage, respectively. Therefore, \citet{pang_clocs_2020} use a much-reduced threshold for proposal generation for each sensor and combine detection candidates before NMS. The final prediction is based on a consistency operation between 2D and 3D proposals in a late fusion fashion.

Other representative examples exploiting attention mechanism for effective fusion of features are proposed by \citet{chen_fast_2019} and \citet{li_3d_2020}.

A totally different approach to combine different inputs is presented by \citet{chen_cooper_2019}. For the specific case of autonomous vehicles, they propose to connect surrounding vehicles and combine their sensor measurements. More specifically, LiDAR data collected from different positions and angels of the connected vehicles are fused together to provide the vehicles with a collective perception of the scene.

In summary, the fusion of different modalities is a vibrant research area within 3DOD. With continuous convolutions and attention mechanism, potential solutions for common issues, such as image-to-point-cloud discrepancies and/or noisy data representations, are already introduced. Nevertheless, fusion approaches still face several unsolved challenges. For example, 2D-driven fusion approaches such as cascaded methods are always constrained by the quality of the 2D detection during the first stage. Therefore, they may fail in cases that can only be observed properly from the 3D space. Feature fusion approaches, on the other hand, generally face the difficulty to fuse different data structures. Consider the example of fusing images and LiDAR data. While images provide a dense, high-resolution structure, LiDAR point clouds show a sparse structure with a comparably low resolution. The workaround of transforming point clouds into another representation inevitably leads to a loss of information. Another challenge for fusion approaches is that crop and resize operations to fuse proposals of different modalities may destroy the feature structure derived from each sensor. Thus, a forced concatenation of a fixed feature vector size could result in imprecise correspondence between the different modalities.

\section{Detection Module}
\label{section:Detection Module}

The detection module depicts the final stage of the pipeline. It uses the extracted features to perform the multi-task consisting of classification, localization along with the bounding box regression and object orientation determination.

Early 3DOD approaches either relied on (i) \textit{template and keypoint matching algorithms}, such as matching 3D CAD models to the scene (Section~\ref{section:Detection Module_Template_Keypoint_matching}) or (ii) suggested handcrafted SVM-classifiers using \textit{sliding window approaches} (Section~\ref{section:Detection Module_Sliding Window}). 

More recent research mainly focuses on \textit{detection frameworks based on deep learning} due to their flexibility and superior performance (Section~\ref{section:Detection Module_Architecture}). Detection techniques of this era can be further divided into (i)~\textit{anchor-based detection} (Section~\ref{section:Detection Module_Anchor-based}), (ii)~\textit{anchorless detection} (Section~\ref{section:Detection Module_Anchorless}) and (iii)~\textit{hybrid detection} (Section~\ref{section:Detection Module_Hybrid}).

\subsection{Template and Keypoint Matching Algorithms}
\label{section:Detection Module_Template_Keypoint_matching}

A natural approach to classifying objects is to compare and match them against a template database. These approaches typically leverage 3D CAD models to synthesize object templates that guide geometric reasoning during inference. Applied matching algorithms use parts or whole CAD models of the objects to classify the candidates.

\citet{teng_surface-based_2014} follow a surface identification approach. Therefore, they accumulate a surface object database of RGB-D images taken from different viewpoints. Then, the specific 3D surface segment obtained by segmenting the current scene is matched with the surface segments in the database. Key points between the matched surface and the observed surface are then matched for pose estimation.

\citet{crivellaro_novel_2015} initially perform a part detection. For each part, seven so-called 3D control points are projected to represent the pose of the object. Finally, a bounding box matching the constraints on the part and the control points is estimated from a small set of learned objects. 

\citet{kehl_deep_2016} create a codebook of local RGB-D patches from synthetic CAD models. These patches, consisting of a variety of different views, are matched with feature descriptors from the scene to classify the object.

Another matching approach is designed by \citet{he_3d_2017} extending LINE-MOD \citep{hinterstoisser_multimodal_2011}, which combines surface normal orientations from depth images and silhouette gradient orientations from RGB images to represent object templates. LINE-MOD is first used to produce initial detection results based on lookup tables for similarity matching. To exclude the many false positive and duplicate detections, \citet{he_3d_2017} cluster templates that matched with a similar spatial location and only then score the matchings.

Further, \citet{yamazaki_discovering_2018} applied a template matching to point cloud projections. The key novelty is the use of constraints imposed by the spatial relationship between image projection directions which are linked through the shared point cloud. This allows to achieve a consistency of the object throughout the multi-viewpoint images, even in cluttered scenes. 

Another approach is proposed by \citet{barabanau_monocular_2020}. The authors introduce a compound solution of key point and template matching. They observe that depth estimation on monocular 3DOD is naturally ill-posed. For that reason, they propose to use sparse but salient key point features. They initially regress 2D key points and then match them with 3D CAD models to predict object dimension and orientation.

\subsection{Sliding Window Approaches}
\label{section:Detection Module_Sliding Window}

The sliding window technique was largely adopted from 2DOD to 3DOD. Here, an object detector slides in the form of a specified window over the feature map and directly classifies each window position. For 3DOD pipelines, this idea is extended by replacing the 2D window with a spatial rectangular box that slides through a discretized 3D space. However, tested solutions have revealed that traversing a window over the entire 3D space is a very exhaustive task, leading to heavy computations and long inference times. 

One of the popular pioneers in this area were \citet{song_sliding_2014} with their Sliding Shapes approach. They use prior trained SVM classifiers to run exhaustively over a voxelized 3D space.

Similarly, \citet{ren_three-dimensional_2016, ren_3d_2018, ren_clouds_2020} use in all of their works a sliding window approach and extensively leverage pre-trained SVMs. In COG 1.0, for example, they use SVMs along with a cascaded classification framework to learn contextual relationships among objects in the scene. Therefore, they train SVMs for each object category with handcrafted features such as surface orientation. Furthermore, they integrate a Manhattan space layout, which assumes an orthogonal space structure to estimate walls, ceilings and floors for a more holistic understanding of the 3D scene and to restrict the detection. Finally, the contextual information is used in a Markov random field representation problem to consider object relationship in detection \citep{ren_three-dimensional_2016}. 

In the successor model, namely LSS, \citet{ren_3d_2018} observe that the height of the support surface is the primary cause of style variation for many object categories. Therefore, they add support surfaces as a latent part for each object, which they use in combination with an SVM and additionally constraints from the predecessor. 

Even in their latest work, COG 2.0, \citet{ren_clouds_2020} still apply an exhaustive sliding window search for 3DOD, laying their focus on robust feature extraction rather than detection techniques. 

Similarly, \citet{liu_faster_2018} use SVMs learned for each object class based on the feature selection proposed by \citet{ren_three-dimensional_2016}. Through a pruning of candidates by comparing the cuboid size of the bounding boxes with the distribution of the physical size of the objects, they further reduce inference time of detection. 

However, an exhaustive sliding window approach tends to be computationally expensive, since the third dimension significantly increases the search space. Therefore, \citet{wang_voting_2015} exploit the sparsity of 3D representations by adding a voting scheme that is activated only for occupied cells, reducing the computational complexity while preserving mathematical equivalence. Whereas the sliding window approach of \citet{song_sliding_2014} operates linear to the total number of cells in 3D grids, the voting approach by \citet{wang_voting_2015} reduces the operations exclusively to the occupied cells. The voting scheme is explained in more detail in Section \ref{section:Feature Extraction_Voting}.

\citet{engelcke_vote3deep_2017} tie on the success of \cite{wang_voting_2015} and propose to exploit feature-centric voting to detect objects in point clouds in even deeper networks to boost performance.

\subsection{Detection Frameworks based on Deep Learning}
\label{section:Detection Module_Architecture}

All of the above detection techniques, which are solid solutions for their specific use cases, are based on manually developed features and are difficult to transfer. Thus, to exploit more robust features and improve detection performance, most modern detection approaches are based on deep learning models.

As with 2DOD, detection networks for 3DOD relying on deep learning can be basically grouped into two meta frameworks: (i)~\textit{two-stage detection frameworks} (Section~\ref{section:Detection Module_Two-stage}) and (ii)~\textit{single-stage detection frameworks} (Section~\ref{section:Detection Module_single stage}).

To provide a basic understanding of these two concepts, we will briefly revisit the major developments for 2D detection frameworks in the following subsections.

\subsubsection{Two-Stage Detection Frameworks}
\label{section:Detection Module_Two-stage}

As the name indicates, two-stage frameworks perform the object detection task in two stages. In the first stage, spatial sub-regions of the input image are identified that contain object candidates, commonly known as region proposals. The proposed regions are coarse predictions that are scored based on their “objectness”. Regions with a high probability of containing an object will achieve a high score and are used as input to the second stage. These unrefined predictions often lack localization precision. Therefore, the second stage mainly improves the spatial estimation of the object through a more fine-grained feature extraction. The following multi-task head then outputs the final bounding box estimation and classification score.

A seminal work following this central idea is that of \citet{girshick_rich_2014}, who introduced \textit{region-based CNN} (R-CNN). Instead of dealing with a huge amount of region proposals via an exhaustive sliding window procedure, R-CNN integrates the selective search algorithm \citep{uijlings_selective_2013} to extract just about 2,000 category-independent candidates. More specifically, selective search is based on a hierarchical segmentation approach which recursively combines smaller regions into larger ones based on similarity in color, texture, size and fill. Subsequently, the 2,000 generated region proposals are cropped and warped into fixed size images in order to be fed into a pre-trained and fine-tuned CNN. The CNN acts as feature extractor to produce a feature vector with a fixed length, which can then be consumed by binary SVM classifiers trained independently for each object class. At the same time, the CNN features are used for the class-specific bounding box regression.

The original R-CNN framework proved to be time and memory consuming due to a lack of shared computations between each training step (i.e., CNN, SVM classifiers, bounding box regressors). To this end, \citet{girshick_fast_2015} developed an extension, called \textit{Fast R-CNN}, in which the individual computations were integrated into a jointly trained framework. Instead of feeding the region proposals generated by selective search to the CNN, the two operations are swapped, so that the entire input image is now processed by the CNN to produce a joint convolutional feature map. The region proposals are then projected onto the joint feature map and a fixed-length feature vector is extracted from each region proposal using a region-of-interest pooling layer. Subsequently, the extracted features are consumed by a sequence of fully connected layers to predict the results for the final object classes and the bounding box offset values for refinement purposes \citep{girshick_fast_2015}. This approach saves memory and improves both, the accuracy and efficiency of object detection models \citep{zhao_object_2019, liu_deep_2020}.

Both R-CNN and Fast R-CNN have the disadvantage of relying on external region proposals generated by selective search, which is a time-consuming process. Against this backdrop, \cite{ren_faster_2017} introduced another extension, called \textit{Faster R-CNN}. As an innovative enrichment, the detection framework consists of a \textit{region proposal network} (RPN) as a sub-network for nominating regions of interest. The RPN is a CNN by itself and replaces the functionality of the selective search algorithm. To classify objects, the RPN is connected to a Fast R-CNN model with which it shares convolutional layers and the resulting feature maps.

It initializes multiple reference boxes, called \textit{anchors}, with different sizes and aspect ratios at each possible feature map position. These anchors are then mapped to a lower dimensional vector, which is used for "objectness" classification and bounding box regression via fully connected layers. These are in turn passed to the Fast R-CNN for bounding box classification and fine tuning. Due to the convolutional layers used simultaneously by the RPN and the Fast R-CNN, the architecture provides a highly efficient solution for region proposals \citep{ren_faster_2017}. Furthermore, since Faster R-CNN is a continuous CNN, the network can be trained end-to-end using backpropagation iteratively and handcrafted features are no longer necessary \citep{zhao_object_2019, liu_deep_2020}.

\subsubsection{Single-Stage Detection Frameworks}
\label{section:Detection Module_single stage}
Single-stage detectors present a simpler network by transforming the input into a structured data representation and employing a CNN to directly estimate bounding box parameters and class scores in a fully convolutional manner. 

Object detectors based on region proposals are computationally intensive and have long inference times, especially on mobile devices with limited memory and computational capacities \citep{redmon_you_2016, liu_deep_2020}. Therefore, single-stage frameworks with significant time advantages have been designed, while having acceptable drawbacks in performance in comparison to the heavyweight two-stage region proposal detectors of the R-CNN family. The speed improvement results from the elimination of bounding box proposals as well as the feature resampling \citep{liu_ssd_2016}. Two popular approaches which launched this development are \textit{YOLO} (you only look once)  \citep{redmon_you_2016} and \textit{SSD} (single shot multibox detector) \citep{liu_ssd_2016}.

The basic idea of YOLO is to divide the original input image into an $S\times S$ grid. Each grid cell is responsible for both, classifying the objects within it and predicting the bounding boxes and their confidence value. However, they use features of the entire input image and not only those of proposed local regions. The use of only a single neural network by omitting the RPN allows YOLO-successor FAST YOLO to run in real-time at up to 155 frames per second. However, YOLO exhibits disadvantages in form of comparably lower quality results such as more frequent localization errors, especially for smaller objects \citep{redmon_you_2016}.

The SSD framework also has real-time capability but does not suffer from such severe performance losses as YOLO. Similarly, the model consists of a single continuous CNN but uses the idea of anchors from RPN. Instead of fixed grids as in YOLO, anchor boxes of various sizes are used to determine the bounding boxes. To detect objects of different sizes, the predictions of several generated feature maps of descending resolution are combined. In this process, the front layers of the SSD network are increasingly used for classification due to their size, and the back layers are used for detection \citep{liu_ssd_2016}.

Thanks to regressing bounding boxes and class scores in one stage, single-stage networks are faster than two-stage frameworks. However, features are not learned from predicted bounding-box proposals but from predefined anchors. Hence, resulting predictions are usually not as accurate as those from two-stage frameworks.

Compared to single-stage approaches, proposal-based models can leverage finer spatial information in the second stage, by only focusing on the narrowed-down region of interest, predicted by the first stage. Features get re-extracted for each proposal which achieves more accurate localization and classification, but in turn, increases the computational costs.

The single- and two-stage paradigm can be transferred from 2DOD to 3DOD. Other than that, we want to further distinguish between the detection techniques described in the following.

\subsection{Anchor-based Detection}
\label{section:Detection Module_Anchor-based}

Many modern object detectors make use of anchor boxes, which serve as the initial guess for the bounding box prediction. The main idea behind anchor boxes is to define a certain set of boxes with different scales and ratios that are mapped densely across the image. This exhaustive selection should be able to capture all relevant objects. The boxes that best contain and match the objects are finally retained.

\begin{figure}[htbp]
\centering
\includegraphics[width=0.5\columnwidth]{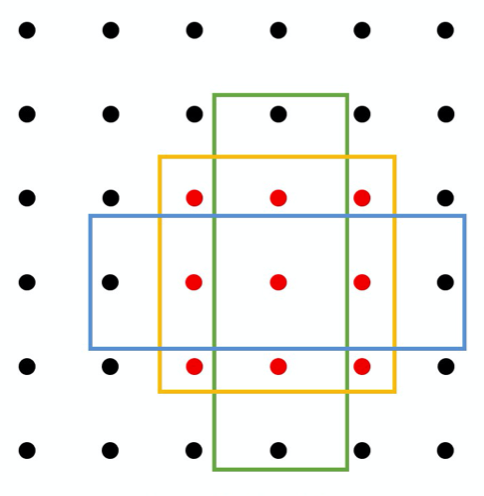}
\caption[Anchor Boxes] {Visualization of several anchors of different scales and ratios centering on the same feature point \citep{Li_2019_anchor_image_ICCV}\label{fig:Anchors}}
\end{figure}

Anchor boxes are boxes of predefined width and length. Both depict important hyperparameters to choose since they must match those of the objects in the dataset. To consider all variation of ratios and scales, it is common to choose a collection of anchor boxes in multiple sizes (see Figure \ref{fig:Anchors}). 

Anchor boxes are proposed in high numbers across the image. Typically, they are initialized at the center of each cell of the final feature map after the feature extraction stage. For localization and classification, a network uses these anchor boxes to learn to predict the offsets between the anchors and the ground truth. Typically, by a combination of the classification confidence of each box and its overlap with the ground truth box, called intersection-over-union (IoU), it is chosen which anchor boxes are discarded and which are kept for refinement purposes \citep{zhong_anchor_2020}.

In Faster R-CNN, three aspect ratios and three scales are used by default, resulting in nine anchors per location. The approach significantly reduces the number of anchors in comparison to existing solutions. This is of particular importance as it enables a computationally acceptable integration of region proposals into the huge search space of 3DOD candidates \citep{song_deep_2016}.

\subsubsection{Two-Stage Detection: 2D Anchors}
\label{section:Detection Module_2Stage_2D_Anchor}

For 2D representations of the spatial space (e.g., BEV, RV and RGB-D), the previously described R-CNN frameworks can be directly implemented and are therefore widely used. Most models first predict 2D candidates from monocular representations. These predictions are then given to a subsequent network that transforms these proposals into 3D bounding boxes by applying various constraints such as depth estimation networks, geometrical constraints or 3D template matching (cf. Section \ref{section:Feature Extraction_Deep Learning_Informed Mono}). 

As a representative example of a 2D anchor-based detection approach in 3DOD, \citet{chen_multi-view_2017} use a 2D RPN at considerably sparse and low-resolution input data such as FV or BEV projections. As this data may not have enough information for proposal generation, \citet{chen_multi-view_2017} assign four 2D anchors - per frame and class - to every pixel in BEV, RV and image feature map, and combine these crops in a deep fusion scheme. The 2D anchors are derived from representative 3D boxes which were obtained by clustering the ground truth objects in the training set by the size and restricting the orientation to 0° and 90°. Leveraging sparsity, they only compute non-empty anchors of the last convolution feature map.

For 3DOD, the 2D anchors are then reprojected to their original spatial dimensionality, which was derived from the ground truth boxes. Following, these 3D proposals serve as the ultimate refinement regression of the bounding boxes.

Further examples are proposed by \citet{deng_amodal_2017},  \citet{zeng_rt3d_2018}, \citet{maisano_reducing_2018} and \citet{beltran_birdnet:_2018}.

\subsubsection{Two-Stage Detection: 3D Anchors}
\label{section:Detection Module_2Stage_3D_Anchor}

Offering a relatively fast detection, 2D anchor-based detectors are not as suitable for high-precision detection. Therefore, a growing part of research is devoted o new and more complex 3D anchor-based detection.

An initial attempt to deploy region proposals with 3D anchors was made by \citet{chen_3d_2015} when introducing 3DOP based on handcrafted features and priors. 3DOP uses depth features in RGB-D point clouds to score candidate boxes in spatial space. A little later, \citet{song_deep_2016} exploit more powerful deep learning features for candidate creation in their seminal work of Deep Sliding Shapes. Inspired by Faster R-CNN, Deep Sliding Shapes divides the 3D scene, obtained from RGB-D data, into voxels and then designs a 3D convolutional RPN to learn objectness for spatial region proposals. The authors define the anchor boxes for each class based on statistics. For each anchor with non-square ground planes, they define an additional anchor with the same size but rotated by 90°. This results in a set of 19 anchors for their indoor scenarios. Given their experiments on the SUN RGB-D \citep{Song_2015_CVPR} and NYUv2 \citep{Silberman:ECCV12} datasets, the total number of anchors per image is about 1.4 million, in comparison to 2,000 anchors per RGB image frame through the selective search algorithm in an R-CNN. Thus, the huge number of anchors leads to extreme computation costs. 

Apart from regular spatial representations in the form of a voxelized spatial space, \citet{xu_pointfusion:_2018} leverage a point-wise representation for anchor-based detection. The input 3D points are used as dense spatial anchors and a prediction is performed on each of the points with two connected MLPs. Similarly, \citet{yang_ipod_2018} define two anchors to propose a total of six 3D candidates on each point of the point cloud. To reduce the number of proposals, they apply a 2D semantic segmentation network which is mapped to the 3D space and eliminates all proposals made on background points. Subsequently, the proposals are refined and scored by a lightweight PointNet prediction network.

\subsubsection{Single-Stage Detection: 2D Anchors}
\label{section:Detection Module_1Stage_2D_Anchor}

Similar to two-stage architectures, the 2D anchor-based single-stage detector framework can be directly applied to 2D-based representations. Exemplary representatives can be found in the work from \citet{liang_deep_2018}, \citet{meyer_sensor_2019}, \citet{ali_yolo3d_2019} and \citet{he_structure_2020}.

\citet{ali_yolo3d_2019}, for instance, use the average box dimensions for each object class from the ground truth dataset as 3D reference boxes and derive 2D anchors. Then a single-stage YOLO framework is applied to a BEV representation and two regression branches are added to produce the \textit{z}-coordinate of the center of the proposal as well as the height of the box.

To enhance prediction quality, \citet{he_structure_2020} perform the auxiliary detection task of point-wise foreground segmentation prior to exploiting anchor-based detection. Subsequently, they estimate the object center with a 3D CNN and only then reshape the feature maps to BEV and employ anchor-based 2D detection.

For further improvement, \citet{gustafsson_accurate_2021} design a differentiable pooling operator for 3D to extend the SA-SSD approach of \citet{he_structure_2020} by a conditional energy-based regression approach instead of the commonly used Gaussian model.

\subsubsection{Single-Stage Detection: 3D Anchors}
\label{section:Detection Module_1Stage_3D_Anchor}

Single-stage detection networks that use anchors based on 3D representations are particularly focuses on extracting meaningful and rich features in the first place, since 3D detection methods are not yet as mature as their equivalent 2D ones. Likewise, the missing performance boost due to the lack of an additional refinement stage must be compensated.

As an exemplary approach, \citet{zhou_voxelnet_2018} introduce the seminal VFE module as an approach for discriminative feature extraction (cf. Section~\ref{section:Feature Extraction_Feature Initalization}). As of today, it is the state-of-the-art encoding for voxel-wise detection models. Having access to these meaningful features, they only use a simple convolutional middle layer in combination with a slightly modified RPN for a single-stage detection purpose.

Observing the problem of high inference times of 3D CNNs in volumetric representations, \citet{sun_3d_2018} introduce a single-stage 3D CNN, treating detection and recognition as one regression problem in a direct manner. Therefore, they develop a deep hierarchical fusion network capturing rich contextual information.

Further exemplary representatives for single-stage 3D anchor detection are proposed by \cite{yan_second_2018} or \cite{li_3d_guivant_2019}, which mainly build upon the success of VoxelNet.

\subsection{Anchorless Detection}
\label{section:Detection Module_Anchorless}

Anchorless detection methods are commonly based on point- or segment-wise detection estimates. Instead of generating candidates, the whole scene is densely classified, and the individual objects and their respective position are derived directly. Apart from a larger group of approaches that use \textit{fully convolutional networks (FCNs)} (Section~\ref{section:Detection Module_FCN}), there exist several \textit{other individual solutions} (Section~\ref{section:Detecton Module_Anchorless - Others}) that propose anchor-free detection models.

\subsubsection{Approaches Based on Fully Convolutional Networks}
\label{section:Detection Module_FCN}

Rather than exploiting anchor-based region proposal networks,  \citet{li_vehicle_2016} pioneered the idea of extending fully convolutional networks (FCNs) \citep{long_fully_2015} to 3DOD. The proposed 3D FCN does not require candidate regions for detection, but implicitly predicts objectness over the entire image. Instead of generating multiple anchors over the feature map, the bounding box then gets directly determined over the objectness regions. \citet{li_3d_2017} further extend this approach in their successor model by going from depth map data to a spatial volumetric representation derived from a LiDAR point cloud.

\citet{kim_lidar_2017} use a two-stage approach, initially predicting candidates in a projection representation based on edge filtering. Leveraging edge detection, objects get segmented and unique box proposals are generated based on the edge boundaries. In the second stage, the authors then apply a region-based FCN to the region of interest.

\citet{meyer_sensor_2019, meyer_lasernet_2019} both employ a mean shift clustering for detection. They use an FCN to predict a distribution over 3D boxes for each point of the feature map independently. In conclusion, points on the same object should predict a similar distribution. To eliminate the natural noise of the prediction, they combine per-point prediction through mean shift clustering. Since all distributions are class-dependent and multimodal, the mean shift has to be performed for each class and modality separately. For efficiency reasons, mean shift clustering is performed over box centers instead of box corners, thereby reducing dimensionality. 

Further representatives using FCNs are \citet{yang_hdnet_2018, yang_pixor_2018}, who use hierarchical multi-scale feature maps, and \cite{wang_frustum_2019-1}, who apply a sliding-window-wise application of an FCN. These networks output pixel-wise predictions at a single stage, with each prediction corresponding to a 3D object estimate.

\subsubsection{Other Approaches}
\label{section:Detecton Module_Anchorless - Others}

Since point-based representations do not admit convolutional networks, models processing the raw point cloud need to find other solutions to apply detection mechanisms. In the following, we summarize some of the innovative developments.

A seminal work that offers a pipeline for directly working on raw point clouds was proposed by \citet{qi_deep_2019} when introducing \textit{VoteNet}. The approach integrates the synergies of 3D deep learning models for feature learning, namely PointNet++ \citep{qi_pointnet++_2017-1} and Hough Voting \citep{leibe_robust_2008}. Since the centroid of a 3D bounding box is most likely far from any surface point, the estimation of bounding box parameters that are based solely on point clouds is a difficult task. By considering a voting mechanism, the authors generate new points that are located close to object centroids, which are used to produce spatial location proposals for the corresponding bounding box. They argue that a voting-based detection is more compatible with sparse point sets as compared to RPNs since RPNs have to carry out extra computations to adjust the bounding box without having an explicit object center. Furthermore, the center is likely to be in an empty space of the point cloud. 

\begin{figure*}[htbp!]
\centering
\includegraphics[width=1\textwidth]{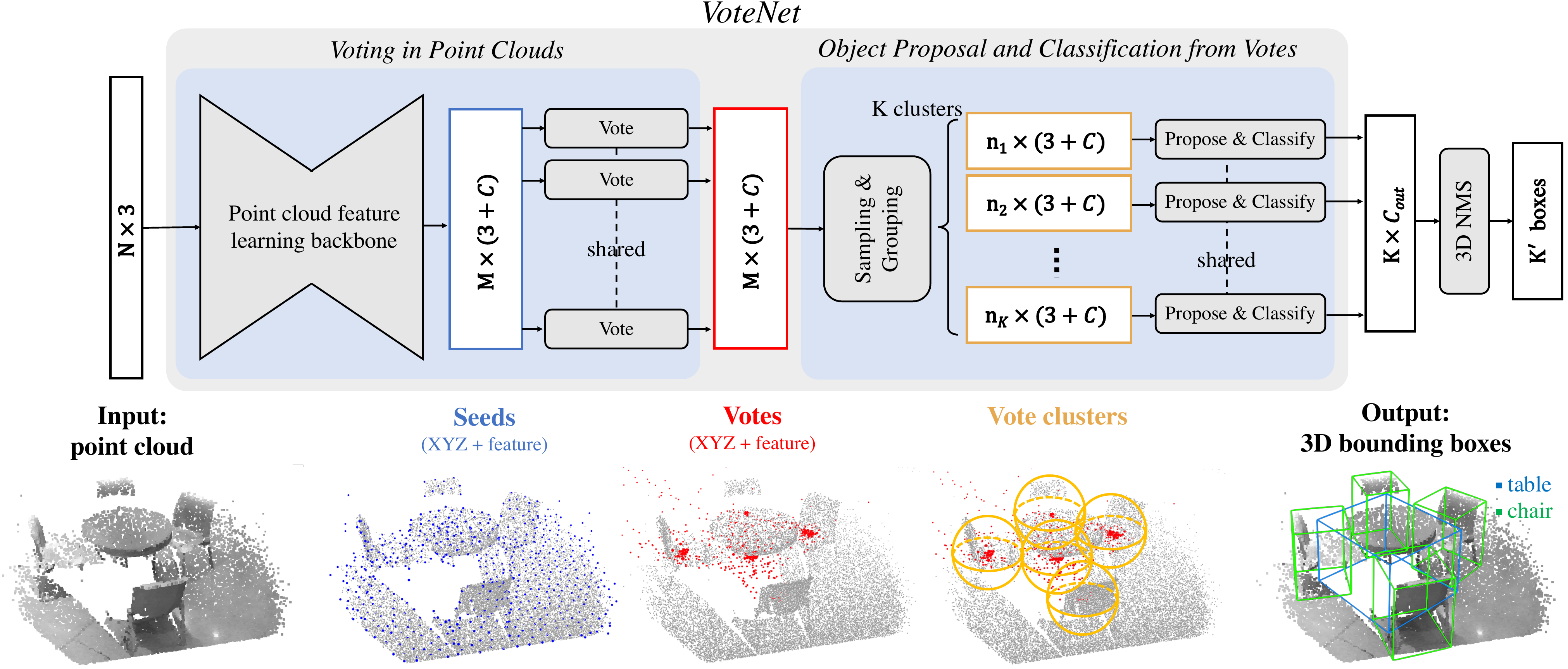}
\caption[Illustration of the architecture and the steps performed by VoteNet] {Illustration of the architecture and the steps performed by VoteNet \citep{qi_deep_2019}\label{fig:VoteNet}}
\end{figure*}

Figure~\ref{fig:VoteNet} illustrates the approach. First, a backbone network based on PointNet++ is used to learn features on the points and derive a subset of points (seeds). Each seed proposes a vote for the centroid by using Hough voting (votes). The votes are then grouped and processed by a proposal module to provide refined proposals (vote clusters). Eventually, the vote clusters are classified and bounding boxes are regressed.

VoteNet provides accurate detection results even though it relies solely on geometric information. To further enhance this approach, \citet{qi_imvotenet_2020} propose ImVoteNet. The successor complements VoteNet by utilizing the high resolution and rich texture of images to fuse 3D votes in point clouds with 2D votes in images.

Another voting approach is proposed by \citet{yang_3dssd_2020}. The authors first use a voting scheme similar to \citet{qi_deep_2019} to generate candidate points as representatives. The candidate points are then further treated as object centers and the surrounding points are gathered to feed an anchor-free regression head for bounding box prediction. 

\citet{pamplona_pointnet_2019} propose an on-road object detection method where they eliminate the ground plane points and then establish an occupation grid representation. The bounding boxes are then extracted for occupation regions that contain a certain number of points. For classification purposes, PointNet is applied.

\citet{shi_pointrcnn_2019} use PointNet++ for a foreground point segmentation. This contextual dense representation is further used for a bin-based 3D box regression and then refined through point cloud region pooling and a canonical transformation. 

Besides their anchor-based solution, \citet{shi_points_2020} also conduct experiments on an anchor-free solution, reusing the detection head of PointRCNN \citep{shi_pointrcnn_2019}. They examined that while the anchorless application is more memory efficient, the anchor-based strategy results in a higher object recall. 

Similar to \citet{shi_pointrcnn_2019}, \citet{li_3d_2020} also exploit a foreground segmentation of the point cloud. For each foreground point, an IoU-sensitive proposal is produced, which leverages the attention mechanism. This is done by only taking the most relevant features into account as well as further geometrical information about the surroundings. For the final prediction, they add a supplementary IoU-perception branch to the commonly used classification and bounding box regression branch for a more accurate instance localization. 

Other than that, \citet{zhou_joint_2020-1} introduce spatial embedding-based object proposals. A point-wise semantic segmentation of the scene is used in combination with a spatial embedding for instance segmentation. The embedding consists of an assembling of all foreground points into their corresponding object centers. After clustering, a mean bounding box is derived for each instance that is again further refined by a network based on PointNet++. 

\subsection{Hybrid Detection}
\label{section:Detection Module_Hybrid}

Next to representation and feature extraction fusion (cf. Section~\ref{section:Fusion}), there are also approaches for fusing detection modules. Two-stage detection frameworks, especially representation-fusion-driven models, generally prefer to exploit 3D detection methods such as anchor-based 3D CNNs, 3D FCNs or PointNet-like architectures for the refinement stage after an originally lightweight 2D-based estimation was performed in the first place. This opens the advantage of a precise prediction using the spatial space through 3D detection frameworks, which are otherwise too time-consuming to be applied to the entire scene. In the following, we classify models that use multiple detection techniques as hybrid detection modules.

Exemplary, \citet{wang_frustum_2019-1} apply multiple methods to finally give a prediction. First, they use an anchor-based 2D detection for proposal generation, which are then extruded as frustums into 3D space. Thereafter, an anchorless FCN detection technique is applied to classify the frustum in a sliding window fashion along the frustum axis to output the ultimate prediction.

A frequently exercised approach is to extrude 2D proposals into the spatial domain of point clouds. The pioneer for this technique is Frustum-PointNet \citep{qi_frustum_2018}, enabling PointNet for the task of object detection. Since PointNet can effectively segment the scene, but not produce location estimations, the authors use preceding 2D anchor-based proposals, which then are classified by PointNet.

Likewise, \citet{ferguson_2d-3d_2019} as well as \citet{shen_frustum_2020} propose a quite similar idea. They first reduce the search space extruding a frustum from a 2D region proposal into 3D space and then use a 3D CNN for the detection task.

Apart from extruding frustums of the initial 2D candidates, proposals in projection representations are often converted into fully specified 3D proposals in space. This is possible because the projection occupies depth information enabling the mapping of the 2D representation to 3D. 

For example, \citet{zhou_fvnet_2019}, \citet{shi_pv-rcnn_2020} and \citet{deng_voxel_2021} transfer the proposals of the 2D-representation into a spatial representation not by extruding an unrestricted search space along the $z$-axis but as fully defined anchor-based 3D proposals by a 2D to 3D conversion of the projections. 3D-compatible detection methods such as anchorless and anchor-based 3D CNNs or PointNets can then be deployed for refinement.

\citet{chen_fast_2019} first generate voxelized candidate boxes that are further processed in a point-wise representation during a second stage. 3D and 2D CNNs are stacked upon a VFE-encoded representation for proposals in the first stage, and only then a PointNet-based network is applied for the refinement of the proposals.

\citet{ku_monocular_2019} use an anchor-based monocular 2D detection to estimate the spatial centroid of the object. An object instance is then reconstructed and laid on the proposal in the point cloud, helping to regress the final 3D bounding box.

In contrast to anchor-based detection, \citet{leal-gupta_3d_2019} do not execute a dense pixel-wise regression of the bounding box but initially estimates key points in the form of the bottom center of the object. Only those key points and their nearest neighbors are then used to produce a comparatively low number of positive anchors, accelerating the detection process. 

Similar to other techniques, matching algorithms are likewise fused in hybrid frameworks. For example, \citet{chabot_deep_2017} use a network to output 2D bounding boxes, vehicle part coordinates and 3D bounding box dimensions. Then they match the dimensions and parts derived in the first step with CAD templates for final pose estimation. Further, \citet{du_general_2018} score the 3D frustum region proposals by matching them with a predefined selection consisting of three car model templates. Similarly, \citet{wang_frustum_2019-1} use a combination of PointNet and 2D-3D consistency constraints within the frustum to locate and classify the objects.

Instead of commonly fusing detection techniques hierarchically, \citet{pang_clocs_2020} use 2D and 3D anchor-based detection in a parallel fashion to fuse the candidates in an IoU-sensitive way. There is no NMS performed before fusing the proposals because the 2D-3D consistency constraint between the proposals eliminates most elements.

In summary, hybrid detection approaches try to compensate for the inferiorities of a single detection framework. While some of these models reach remarkable results, the harmonization of two different systems represents a major challenge. Next to the advantages also the disadvantage of the specific technology needs to be handled. In the case of a compound solution between a 2D- and PointNet-like technique, the result offers an obvious improvement in inference speed, as the initial prediction is usually performed in a lightweight 2D detection framework limiting the search space for the PointNet. Yet, the accuracy and precision of detection are less favorable in comparison to full 3D region proposal networks, since the possible uncertainties of a 2D detection are inherited to the hierarchical next step.

\section{Classification of 3D  Object Detection Pipelines}
\label{section:Classification of literature}

In the previous sections, we gave a comprehensive review of different models and methods along the 3DOD pipeline and emphasized representative examples for every stage with their corresponding design options. In the following, we use our proposed pipeline framework from Figure~\ref{fig:3DOD pipeline} (see Section~\ref{section:3DOD-Pipeline}) to classify each 3DOD approach of our literature corpus to derive a thorough systematization.

For better comparability, we distinguish all 3DOD models according to their data representation. That is, we provide separate classification schemes for (i) \textit{monocular models} (Table~\ref{table:3DOD_Conceptualization_Mono}), (ii) \textit{RGB-D front-view-based models} (Table~\ref{table:3DOD_Conceptualization_RGB-D-FV}), (iii) \textit{projection-based models} (Table~\ref{table:3DOD_Conceptualization_Projection}), (iv) \textit{volumetric grid-based models} (Table~\ref{table:3DOD_Conceptualization_Volumetric}), (v) \textit{point-based models} (Table~\ref{table:3DOD_Conceptualization_Pointwise}) and (vi) \textit{fusion-based models} (Table~\ref{table:3DOD_Conceptualization_Fusion}).

For each 3DOD approach, we provide information on the authors, the year and the name, classify the underlying domain and benchmark dataset(s) and categorize the specified design choices along the 3DOD pipeline.

\begin{table*}[htbp]
\centering
\includegraphics[width=1\textwidth]{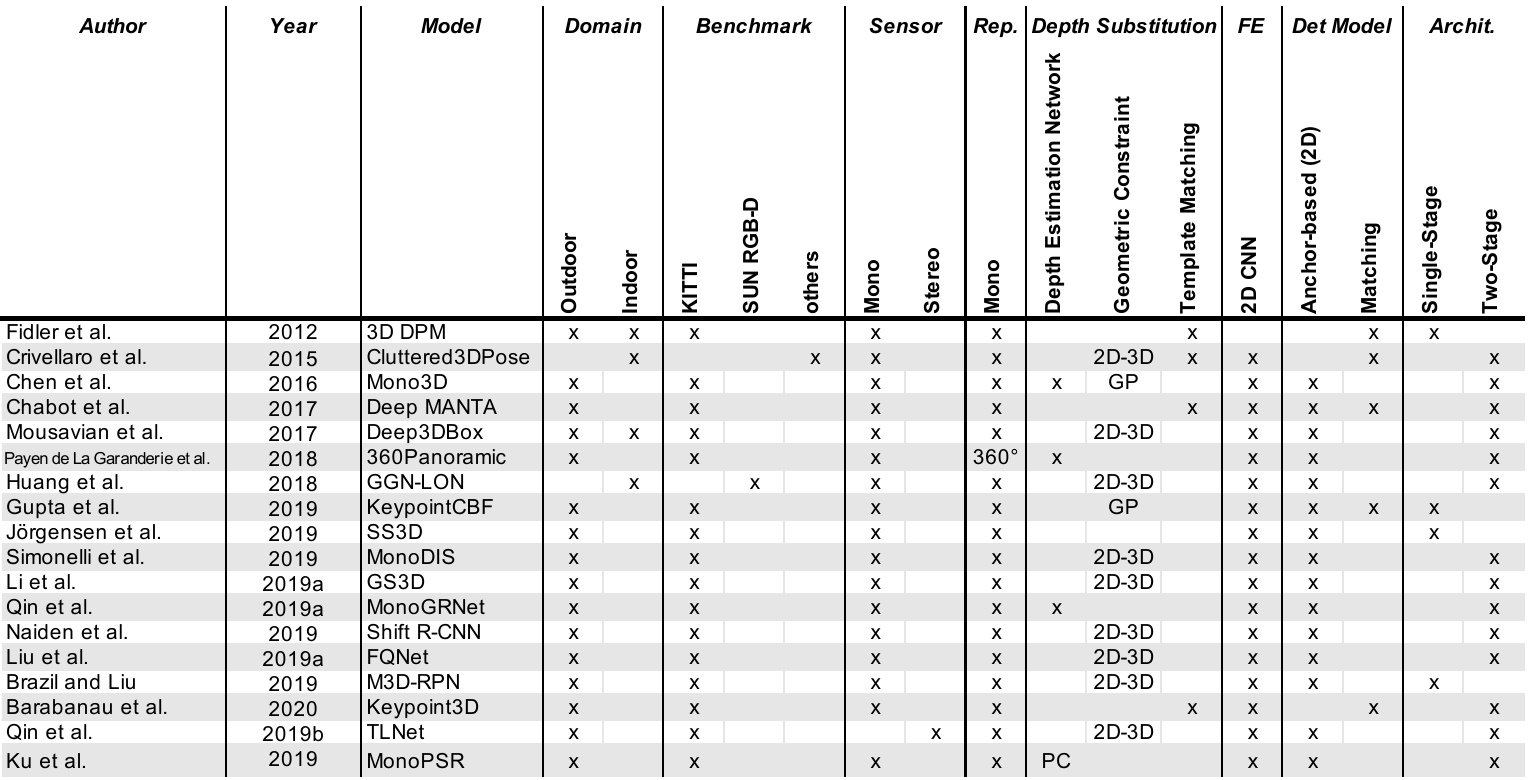}
\caption[Monocular 3DOD model classification] {Classification of monocular 3DOD models (\textit{360°: 360°-monocular image, PC: point cloud, 2D-3D: 2D-3D consistency, GP: groundplane)}
\label{table:3DOD_Conceptualization_Mono}}
\end{table*}

\begin{table*}[htbp]
\centering
\includegraphics[width=0.9\textwidth]{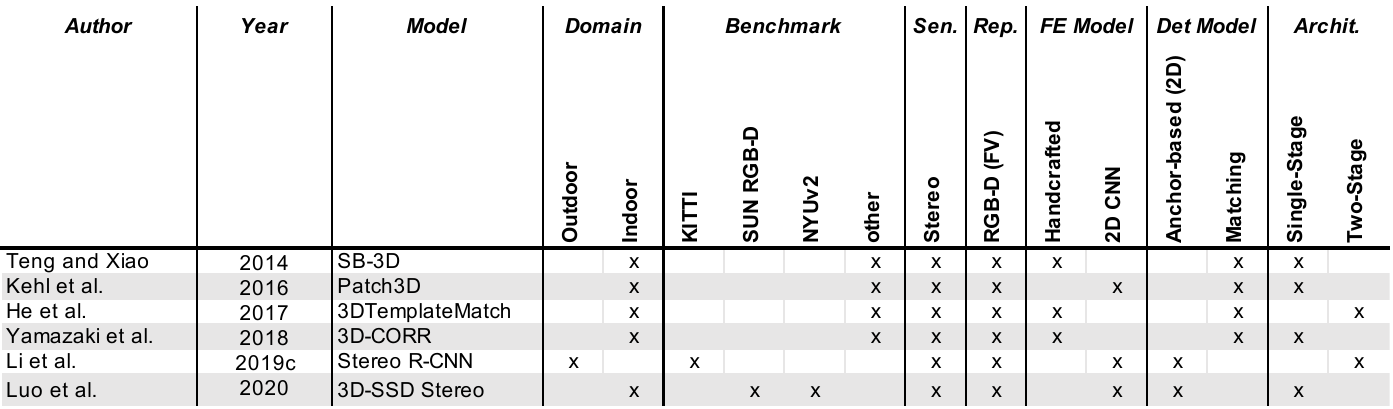}
\caption[RGB-D front-view-based 3DOD model classification] {Classification of RGB-D front view-based 3DOD models
\label{table:3DOD_Conceptualization_RGB-D-FV}}
\end{table*}

\begin{table*}[htbp]
\centering
\includegraphics[width=0.9\textwidth]{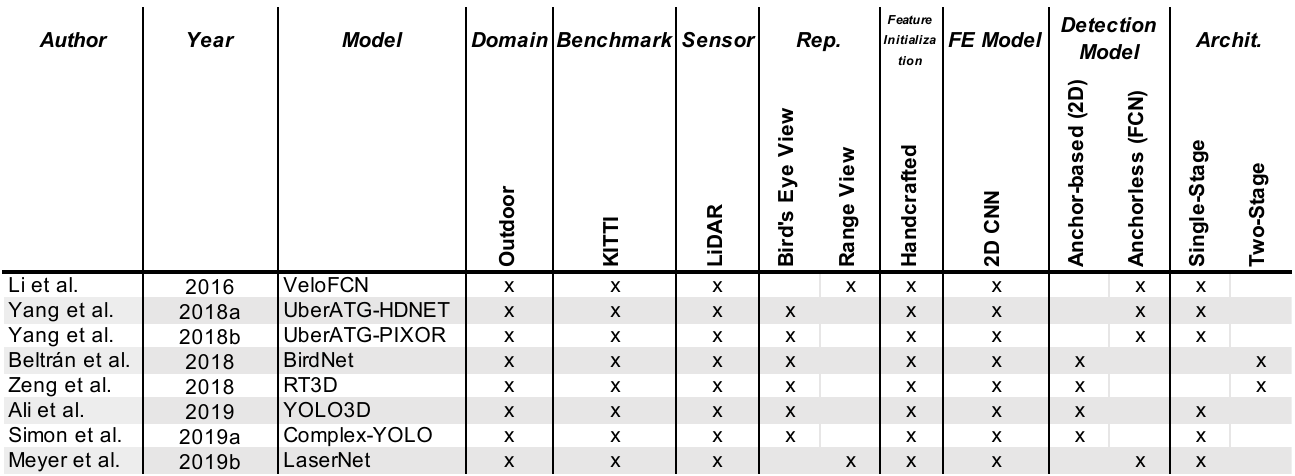}
\caption[Projection-based 3DOD model classification] {Classification of projection-based 3DOD models
\label{table:3DOD_Conceptualization_Projection}}
\end{table*}

\begin{table*}[htbp]
\centering
\includegraphics[width=1\textwidth]{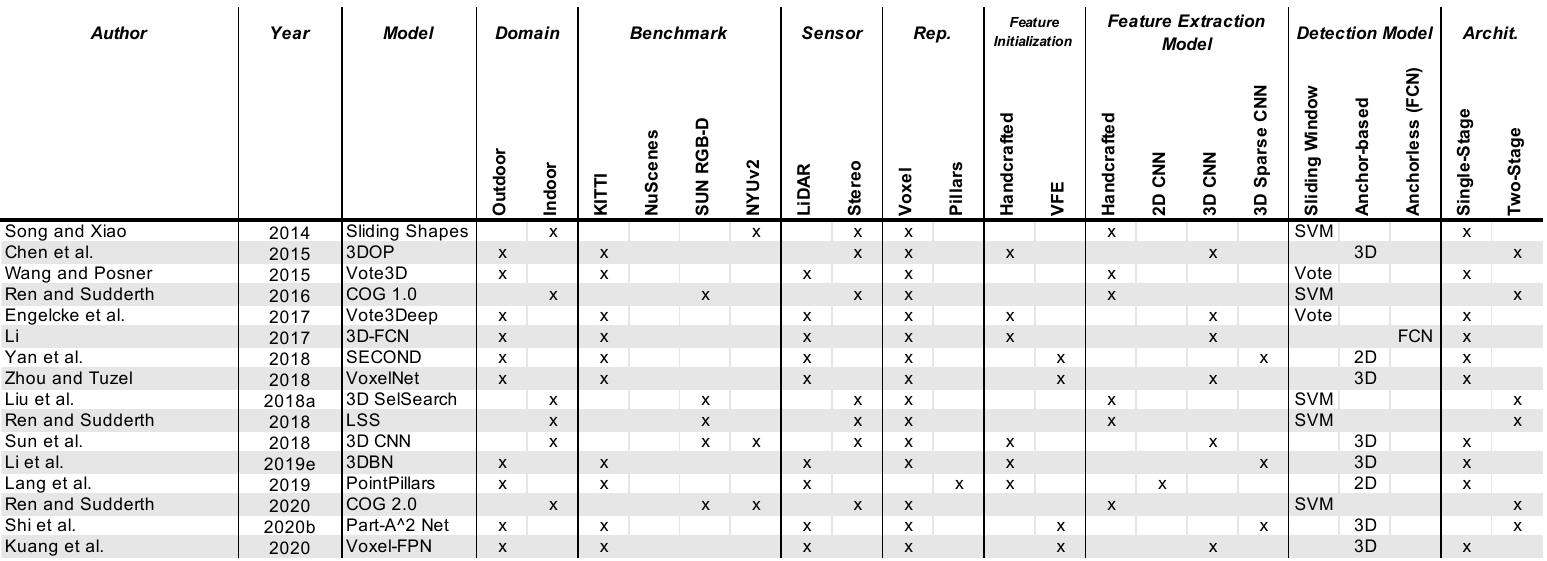}
\caption[Volumetric grid-based 3DOD model classification] {Classification of volumetric grid-based models 3DOD models (\textit{SVM: support vector machine, Vote: voting scheme, FCN: fully convolutional network)}
\label{table:3DOD_Conceptualization_Volumetric}}
\end{table*}

\begin{table*}[htbp]
\centering
\includegraphics[width=0.6\textwidth]{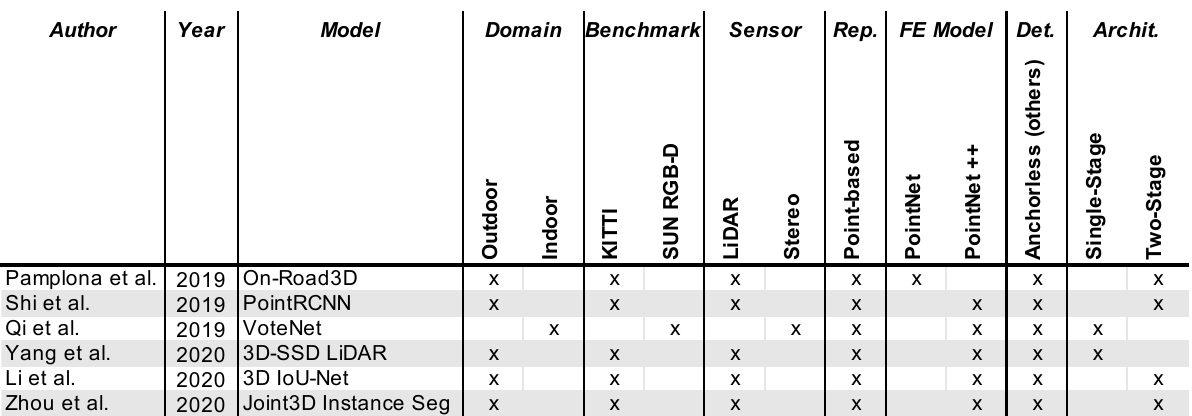}
\caption[Point-based 3DOD model classification] {Classification of point-based 3DOD models\label{table:3DOD_Conceptualization_Pointwise}}
\end{table*}

\begin{table*}[htbp]
\centering
\includegraphics[width=1\textwidth]{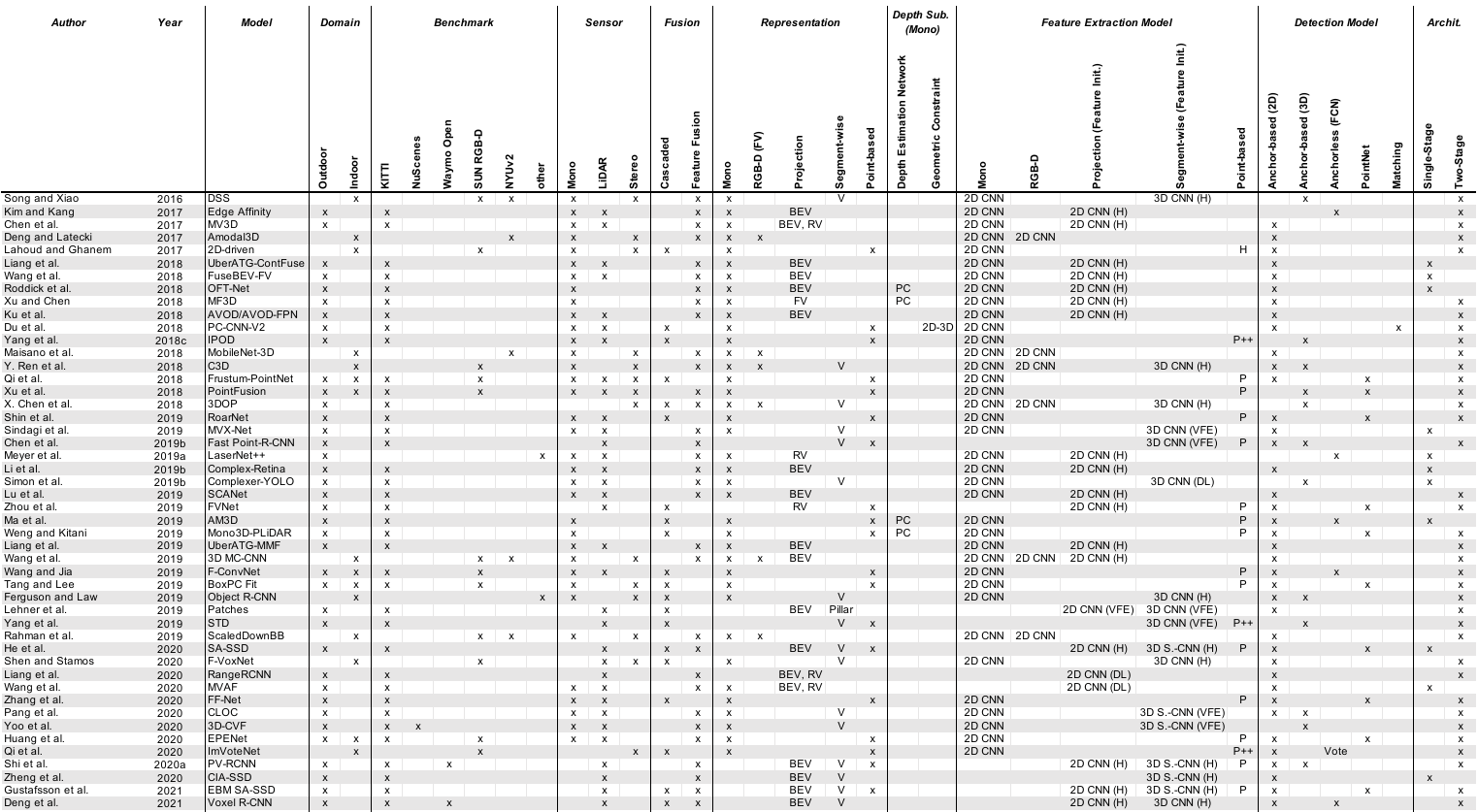}
\caption[Fusion-based 3DOD model classification] {Classification of fusion-based 3DOD models (\textit{BEV: bird's eye view, RV: range view, V: voxel, PC: point cloud, 2D-3D: 2D-3D consistency, H: handcrafted, VFE: voxel feature encoding, DL: deep learning, P: PointNet, P++: PointNet++, 3D S.-CNN: 3D sparse CNN, Vote: voting scheme)}\label{table:3DOD_Conceptualization_Fusion}}
\end{table*}

The resulting classification can help researchers and practitioners alike to get a quick overview of the field, spot developments and trends over time and identify comparable approaches for further development and benchmark purposes. As such, our classification delivers an overview of different design options, provides structured access to knowledge in terms of a 3DOD pipeline catalog and offers a setting to position individual configurations of novel solutions on a more comparable basis.

\section{Concluding Remarks and Outlook}
\label{section:Conclusion}

3D object detection is a vivid research field with a great variety of different approaches. The additional third dimension compared to 2D vision forces the exploration of completely new methods, while mature 2DOD solutions can only be adopted to a limited extent. Hence, new ideas and data usage are emerging to handle the advanced problem of 3DOD resulting in a fastly growing research field that is finely branched in its trends and approaches.

From a broader perspective, we could observe several global trends within the field.
For instance, a general objective of current research is the efficiency optimization of increased computation and memory resource requirements due to the extra dimension of 3DOD, with the ultimate goal of finally reaching real-time detection. 

More recent approaches increasingly focus on fully leveraging point-wise representations, since it promises the best conception of 3D space. As of the literature body of this work, PointNet-based approaches remain the only method so far that can directly process the raw point representation. 

Furthermore, we observe that the fusion of feature extraction and detection techniques as well as data representation are the most popular approaches to challenge common problems of object detection, such as amodal perception, instance variety and noisy data. For feature fusion approaches, the development of attention mechanisms to efficiently fuse features based on their relevance is a major trend. Additionally, the introduction of continuous convolutions facilitates complex modality mapping. In general, hybrid detection models enjoy popularity for exploiting lightweight proposals to restrict the search space for more powerful but heavier refinement techniques.

Summarizing, this work aimed to complement previous surveys, such as those from \citet{arnold_survey_2019}, \citet{guo_deep_review_2020} and \citet{fernandes_point-cloud_2021}, by closing a gap of not only focusing on a single domain and/or specific methods of 3D object detection. Therefore, our search was narrowed only to the extent that the relevant literature should provide a design of an entire pipeline for 3D object detection. We purposely included all available approaches independent of the varieties of data inputs, data representations, feature extraction approaches and detection methods. Therefore, we reviewed an exhaustively searched literature corpus published between 2012 and 2021, including more than 100 approaches from both indoor applications as well as autonomous driving applications. Since these two application areas cover the vast majority of existing literature, our survey may not be subject to the risk of missing major developments and trends.

A particular goal of this survey was to give an overview of all aspects of the 3DOD research field. Therefore, we provided a systematization of 3DOD methods along the model pipeline with a proposed abstraction level that is meant to be neither too coarse nor too specific. As a result, it was possible to classify all models within our literature corpus to structure the field, highlight emerging trends and guide future research. 

At the same time, however, it should be noted that the several stages of the 3DOD pipeline can be designed with a much broader variety and that each stage, therefore, deserves a much closer investigation in subsequent studies. \citet{fernandes_point-cloud_2021}, for instance, go into further details for the feature extraction stage by aiming to organize the entanglement of different extraction paradigms. Yet, we believe that a full conception and systematization of the entire field has not been reached.

In addition, we would like to acknowledge that, in individual cases, it might be difficult to draw a strict boundary for the classification of models regarding design choices and stages. 3D object detection is a highly complex and multi-faceted field of research, and knowledge from 2D and 3D computer vision as well as continuous progress in artificial intelligence and machine learning are getting fused.

Thus, our elaboration of the specific stages marks a broad orientation within the configuration of a 3DOD model and should be rather seen as a collection of possibilities than an ultimate and isolated choice of design options. Especially modern models often jump within these stages and do not follow a linear way along the pipeline, making a strict classification challenging.

Furthermore, as with any review, our work represents only a snapshot in time of the extremely fast-evolving 3DOD research area. Only recently, new deep learning methods have entered the field of 3DOD that can handle point cloud processing, such as graph-based neural networks \cite[e.g.,][]{shi_point-gnn_2020}, kernel point convolutions \cite[e.g.,][]{thomas_kpconv_2019} and Transformer-based networks \cite[e.g.,][]{misra_end--end_2021, mao_voxel_2021, pan_3d_2021}. Likewise, researchers have come up with novel innovations along the 3DOD pipeline, such as adaptive spatial feature aggregation \citep{ji_stereo_2022} or semantical point-voxel feature interaction \citep{wu_pv-rcnn_2022}. These new methods and networks are meant to overcome the limitation of previous architectures. Yet, the potential of existing methods such as PointNet has probably not been reached. A consideration of these fairly new concepts and methods in this survey would have exceeded the scope of this work. However, for future work, a further investigation into these directions could be of great interest. To this end, our proposed systematization offers a great starting point to classify and compare existing as well as emerging approaches on a structured basis.

For future research, we suggest looking into a combination of methods along all stages of the 3DOD pipeline. We recommend examining these aspects independent of the pipeline since the fusion of techniques often occurs in a non-linear way.

Finally, this work could support a practical creation of individual modules or even a whole new 3DOD model, since the systematization along the pipeline can serve as an orientation of design choices within the specific stages.
\section{Acknowledgments}
P.Z. acknowledges funding from the Federal Ministry of Education and Research (BMBF), Germany within the project ‘‘White-Box-AI’’ (grant number 01IS22080). 

\section{Data availability statement}
All data generated or analyzed during this study are included in this published article.

\section{Competing Interests}

The authors have no competing interests to declare that are relevant to the content of this article.
%\clearpage
\newpage
\section*{Abbreviations}
\label{section:Abbreviations}

\begin{table}[ht]
\begin{tabular}{ll}
2DOD  & 2D object detection               \\
3DOD  & 3D object detection               \\
6DoF  & Six degrees of freedom            \\
BEV   & Bird's eye view                   \\
CAD   & Computer-aided design             \\
CNN   & Convolutional neural networks     \\
COG   & Cloud of oriented gradients       \\
DORN  & Deep ordinal regression network   \\
FCN   & Fully convolutional networks      \\
FPS   & Farthest point sampling           \\
FV    & Front view                        \\
HOG   & Histogram of oriented gradients   \\
IoU   & Intersection-over-union           \\
LiDAR & Light Detection and Ranging       \\
LSS   & Latent support surfaces           \\
MLP   & Multi-layer perceptron            \\
NMS   & Non-maximum suppression           \\
R-CNN & Region-based CNN                  \\
RGB-D & RGB-Depth                         \\
ROI   & Regions of interest               \\
RPN   & Region proposal network           \\
RV    & Range view                        \\
SIFT  & Scale-invariant feature transform \\
SSD   & Single shot detection             \\
SVM   & Support vector machine            \\
TOF   & Time-of-Flight                    \\
VFE   & Voxel feature encoding            \\
YOLO  & You Only Look Once                
\end{tabular}
\end{table}

\bibliographystyle{model2-names} 
\bibliography{cas-refs}

\end{document}